\documentclass[sigconf, screen, svgnames]{acmart}

\usepackage{amsmath}
\usepackage{adjustbox}
\usepackage{booktabs} 
 
\usepackage{tabularx}

\usepackage{graphicx,booktabs,multirow,tabularx,makecell}

\renewcommand\arraystretch{1.2}

\usepackage{siunitx}
\usepackage{tikz}
\usetikzlibrary{arrows.meta,
                calc,
                decorations.pathreplacing,
                decorations.text,
                bending,
                fit,
                intersections,
                matrix,
                positioning,
                shapes,
                backgrounds,
                matrix}

\usepackage{pifont}
\newcommand{\cmark}{\ding{51}}
\newcommand{\xmark}{\ding{55}}

\usepackage{ragged2e}

\usepackage{etoolbox}

\usepackage{fancyhdr}

\makeatletter
\fancypagestyle{supplementarystyle}{%
  \fancyhf{}% clear all
  \fancyhead[LE]{\sffamily\footnotesize Extended Supplementary Material}
  \fancyhead[RO]{\sffamily\footnotesize Extended Supplementary Material}
  \fancyhead[RE]{\sffamily\footnotesize \shortauthors}
  \fancyhead[LO]{\sffamily\footnotesize \shorttitle}
  \fancyfoot[C]{\if@ACM@printfolios\footnotesize\thepage\fi}%
  \renewcommand{\headrulewidth}{\z@}%
  \renewcommand{\footrulewidth}{\z@}%
}
\makeatother

\makeatletter
\pretocmd{\@mkauthors}{%
  \begingroup
  \hypersetup{linkcolor=black,urlcolor=black}%
}{}{}
\apptocmd{\@mkauthors}{%
  \endgroup
}{}{}
\makeatother

\makeatletter
\AtBeginDocument{%
  \@namedef{r@TotPages}{{12}{}}%
}
\makeatother

\AtBeginDocument{%
  }

\copyrightyear{2026}
\acmYear{2026}
\setcopyright{cc}
\setcctype{by}
\acmConference[WWW '26]{Proceedings of the ACM Web Conference 2026}{April 13--17, 2026}{Dubai, United Arab Emirates}
\acmBooktitle{Proceedings of the ACM Web Conference 2026 (WWW '26), April 13--17, 2026, Dubai, United Arab Emirates}
\acmDOI{10.1145/3774904.3792677}
\acmISBN{979-8-4007-2307-0/2026/04}
\begin{document}

\title[VL-KGE: Vision--Language Models Meet Knowledge Graph Embeddings]{VL-KGE: Vision--Language Models Meet\\Knowledge Graph Embeddings}

\author{Athanasios Efthymiou}
\email{a.efthymiou@uva.nl}
\orcid{0000-0001-7163-1115}
\affiliation{%
  \institution{University of Amsterdam}
  \city{Amsterdam}
  \country{The Netherlands}
}

\author{Stevan Rudinac}
\email{s.rudinac@uva.nl}
\orcid{0000-0003-1904-8736}
\affiliation{%
  \institution{University of Amsterdam}
  \city{Amsterdam}
  \country{The Netherlands}
}

\author{Monika Kackovic}
\email{m.kackovic@uva.nl}
\orcid{0000-0002-7423-3902}
\affiliation{%
  \institution{University of Amsterdam}
  \city{Amsterdam}
  \country{The Netherlands}
}

\author{Nachoem Wijnberg}
\email{n.m.wijnberg@uva.nl}
\orcid{0000-0001-8070-8719}
\affiliation{%
  \institution{University of Amsterdam}
  \city{Amsterdam}
  \country{The Netherlands}
}

\author{Marcel Worring}
\email{m.worring@uva.nl}
\orcid{0000-0003-4097-4136}
\affiliation{%
  \institution{University of Amsterdam}
  \city{Amsterdam}
  \country{The Netherlands}
}

\renewcommand{\shortauthors}{Athanasios Efthymiou, Stevan Rudinac, Monika Kackovic, Nachoem Wijnberg, \& Marcel Worring}

\begin{abstract}
Real-world multimodal knowledge graphs (MKGs) are inherently heterogeneous, modeling entities that are associated with diverse modalities. Traditional knowledge graph embedding (KGE) methods excel at learning continuous representations of entities and relations, yet they are typically designed for unimodal settings. Recent approaches extend KGE to multimodal settings but remain constrained, often processing modalities in isolation, resulting in weak cross-modal alignment, and relying on simplistic assumptions such as uniform modality availability across entities. Vision--Language Models (VLMs) offer a powerful way to align diverse modalities within a shared embedding space. We propose Vision--Language Knowledge Graph Embeddings (VL-KGE), a framework that integrates cross-modal alignment from VLMs with structured relational modeling to learn unified multimodal representations of knowledge graphs. Experiments on WN9-IMG and two novel fine art MKGs, WikiArt-MKG-v1 and WikiArt-MKG-v2, demonstrate that VL-KGE consistently improves over traditional unimodal and multimodal KGE methods in link prediction tasks. Our results highlight the value of VLMs for multimodal KGE, enabling more robust and structured reasoning over large-scale heterogeneous knowledge~graphs.
\end{abstract}



% \keywords{knowledge graph embeddings, vision-language models, multimodal learning, fine-art analysis}
\maketitle
\newcommand\webconfavailabilityurl{https://doi.org/10.5281/zenodo.18333433}
\ifdefempty{\webconfavailabilityurl}{}{
\begingroup\small\noindent\raggedright\textbf{Resource Availability:}\\
Code and data associated with this paper are publicly available at \url{\webconfavailabilityurl} and \url{https://github.com/thefth/vl-kge}.
\endgroup
}

\begin{figure}[ht]
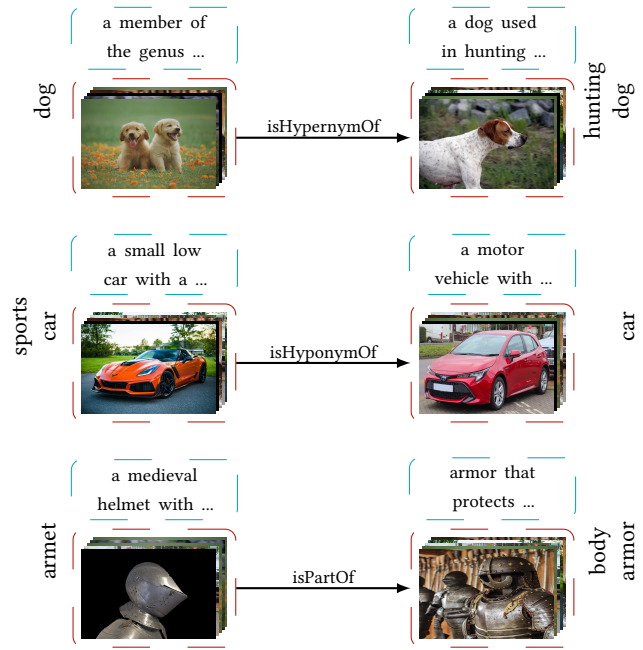

\centering
\input Figures/wn9_img_dataset_task.tex
\caption{Example triples from the WN9-IMG dataset~\cite{wn9_img_dataset}. Entities correspond to ImageNet~\cite{imagenet} synsets, represented by sets of images (shown in \color{red(process)}{red}\color{black}) and their WordNet~\cite{wordnet} textual definitions (shown in \color{cyan(process)}{cyan}\color{black}), connected by semantic relations.}
\label{fig:wn9_img_dataset_task}
\end{figure}

\section{Introduction}\label{sec:introduction}

Knowledge graphs (KGs) are widely used to represent structured knowledge by modeling entities and their relationships across diverse domains, from social networks~\cite{deep_graph_kernels, tu_dataset, graphsage, node2vec, deepwalk} to scientific discovery~\cite{cora, cora_v2, mag_dataset, ogb, ogblsc} and visual arts~\cite{artsagenet, gcnboost, contextnet, graphclip}. To enable effective representation learning on such graphs, knowledge graph embedding (KGE) methods~\cite{transe, rescal, distmult, complex, rotate, nodepiece, recpiece} encode symbolic triples $(h, r, t)$ into continuous vector spaces for knowledge graph completion tasks such as link prediction. However, these methods rely solely on graph structure, overlooking rich multimodal content, such as visual attributes and textual descriptions, that is ubiquitous in many real-world KGs derived from web data.

\begin{figure*}[t]
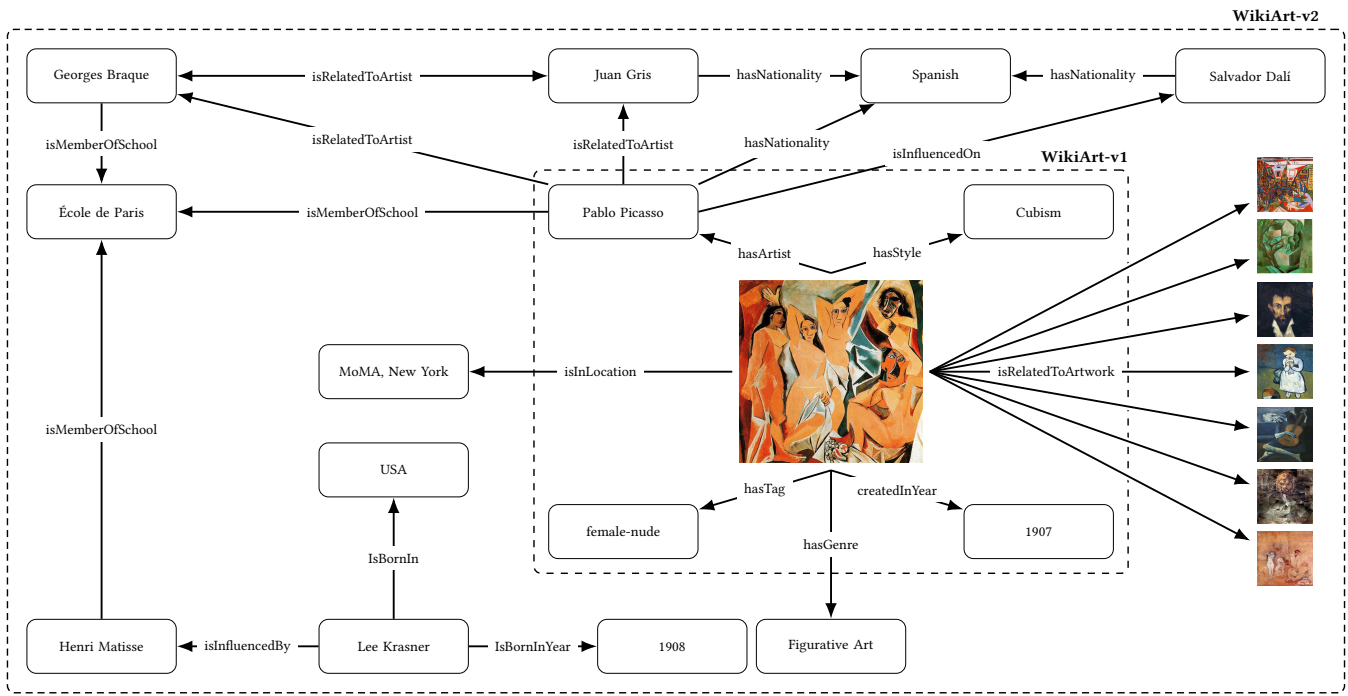

\centering
\input Figures/wikiart_mkg_v1_v2.tex
\caption{Example subgraphs from the WikiArt-MKGs. Artworks are represented visually, while associated entities (e.g., artists, styles, genres, locations) are represented textually. WikiArt-MKG-v1 (inner dashed box) captures core artwork-level relations, whereas WikiArt-MKG-v2 (outer dashed box) extends the graph with additional entity types and richer semantic links.}
\label{fig:wikiart_mkg_v1_v2}
\end{figure*}

Recent work has extended KGE methods to multimodal settings~\cite{transae, dkrl, ikrl, tbkge, mmkrl, otkge}, integrating visual and textual information to enrich entity representations. Yet, existing approaches face two critical limitations. First, they often treat each modality independently, leading to systematic modality misalignment where multimodal representations are not semantically aligned in a shared embedding space. Second, they typically assume that all entities possess all available modalities, an assumption that holds in curated benchmarks, such as WN9-IMG~\cite{wn9_img_dataset} (Figure~\ref{fig:wn9_img_dataset_task}), where each entity is deliberately annotated with structural, visual, and textual attributes. However, this assumption fails in real-world scenarios where modality asymmetry is inherent: different entity types naturally possess different combinations of modalities based on their semantic nature. For instance, in fine art knowledge graphs (Figure~\ref{fig:wikiart_mkg_v1_v2}), artworks are inherently visual objects best represented through images, while artists, stylistic movements, and historical periods are abstract entities characterized primarily through textual descriptions or categorical attributes. Consequently, existing multimodal KGE frameworks struggle to jointly address both modality misalignment and modality asymmetry, thereby significantly limiting their applicability to complex real-world knowledge graphs.

In parallel to making KGE multimodal, Vision--Language Models (VLMs), such as CLIP~\cite{clip}, ALIGN~\cite{align}, and BLIP-2~\cite{blip_2}, have demonstrated strong performance in learning aligned visual and textual representations through contrastive learning or generative pretraining. These models embed multimodal content into a shared semantic space, learning robust and transferable embeddings, making them well-suited to address both modality asymmetry and misalignment. Importantly, VLMs are pretrained on large-scale web data, learning rich associations between images and text that transfer effectively to downstream tasks. The ability of VLMs to provide semantically aligned cross-modal representations without requiring task-specific training opens new opportunities for structured knowledge modeling in multimodal knowledge graphs (MKGs).

In this work, we propose Vision--Language Knowledge Graph Embeddings (VL-KGE), a multimodal KGE framework that integrates pretrained vision-language representations with structural modeling to learn unified representations of entities across modalities. VL-KGE fuses visual and textual representations with structural embeddings into a unified entity representation, while preserving relational semantics. This design explicitly handles modality asymmetry and enables modeling of both intra-modal and cross-modal interactions. To evaluate VL-KGE under both complete and asymmetric modality settings, we first use the standard WN9-IMG benchmark~\cite{wn9_img_dataset}, where all entities have complete modalities. To further assess performance under realistic modality asymmetry, we leverage the WikiArt-v1 dataset~\cite{elgammal2018shape, artsagenet, tan_artgan} and curate a significantly expanded version, WikiArt-v2, with broader coverage of artworks, artists, and detailed metadata on stylistic movements, exhibition locations, and other attributes. From these datasets, we construct two novel fine art multimodal knowledge graphs, namely WikiArt-MKG-v1 and WikiArt-MKG-v2, with the latter incorporating substantially richer artwork-to-attribute, artist-to-attribute, artwork-to-artwork, and artist-to-artist relations. Experimental results demonstrate that VL-KGE consistently improves performance across all benchmarks, with particularly strong gains on the WikiArt-MKGs where modality asymmetry is inherent. Our key contributions are as follows:

\begin{itemize}

\item We propose VL-KGE, a framework that integrates pretrained vision-language representations with structured relational modeling to learn multimodal knowledge graph embeddings.

\item We explicitly address modality asymmetry, enabling VL-KGE to represent heterogeneous entities using only their available modalities and to model intra-modal and cross-modal interactions across visual and textual domains.

\item We introduce WikiArt-v2, WikiArt-MKG-v1, and WikiArt-MKG-v2, substantially expanding fine art datasets to facilitate research on multimodal KGE under modality asymmetry.

\item We demonstrate that VL-KGE consistently improves performance on link prediction tasks across all benchmarks, particularly in modality-asymmetric scenarios.

\end{itemize}

In Section~\ref{sec:related_work}, we review related work on multimodal KGE, vision-language modeling, and computational fine art analysis. In Sections~\ref{sec:approach} and~\ref{sec:experimental_setup}, we detail our proposed VL-KGE framework and experimental setup, respectively. In Section~\ref{sec:results}, we present the results, and in Section~\ref{sec:conclusion}, we conclude with a discussion of our findings.

\section{Related Work}\label{sec:related_work}

Recent advances in vision-language representation learning~\cite{urban_clip, otmkgrl} and knowledge graph embeddings~\cite{relik, ime, fedlu} have enabled new approaches for integrating visual and textual information with structured knowledge~\cite{towards_mkgs_data_spaces, uknow, unigraph_2}. In this section, we review recent work in multimodal knowledge graph embeddings and vision-language models, and discuss computational fine art analysis as an exemplar domain where modality asymmetry is intrinsic.

\subsection{Multimodal Knowledge Graph Embeddings}\label{subsec:multimodal_kge}

Knowledge graph embedding (KGE) methods project entities and relations into continuous vector spaces for knowledge graph completion tasks. Unimodal KGE approaches, such as RESCAL~\cite{rescal}, TransE~\cite{transe}, DistMult~\cite{distmult}, HoLE~\cite{hole}, ComplEx~\cite{complex}, and RotatE~\cite{rotate}, use translation, bilinear, or tensor factorization-based formulations to model relational triples. Subsequent work introduced neural architectures including ConvE~\cite{conve}, ConvKB~\cite{conv_kb}, NodePiece~\cite{nodepiece}, and RecPiece~\cite{recpiece}, which improve parameter efficiency and scalability while preserving structural semantics. However, these methods are inherently unimodal and neglect the rich multimodal context (e.g., visual and textual attributes) present in many real-world knowledge graphs. To address this, early multimodal KGE approaches extended traditional methods with additional feature encoders. IKRL~\cite{ikrl}, TBKGE~\cite{tbkge}, and MMRFAN~\cite{mmrfan} integrate visual features, typically relying on pretrained CNNs for image encoding. DKRL~\cite{dkrl}, MKRL~\cite{mkrl}, and TransAE~\cite{transae} incorporate textual descriptions into entity embeddings, often using CNNs or bag-of-words encoders.

Recent advances have aimed to unify modalities more holistically. MMKRL~\cite{mmkrl} combines visual, textual, and structural signals through adversarial training and component alignment. OTKGE~\cite{otkge} introduces optimal transport to fuse multimodal views. MoSE~\cite{mose} separates and ensembles modality-specific embeddings to reduce cross-modal interference. RSME~\cite{rsme}, MMRNS~\cite{mmrns}, SNAG~\cite{snag}, and MR-MKG~\cite{mr_mkg} further incorporate relation-aware attention, noise masking, and alignment strategies, while models like MTL-KGC~\cite{mtl_kg} and KG-BERT~\cite{kg_bert} explore the integration of pretrained language models for enhanced relational reasoning. LAFA~\cite{lafa} introduces a link-aware mechanism that dynamically weighs the contribution of visual information based on relational context, mitigating noise from irrelevant images. MCKGC~\cite{mckgc} embeds multimodal information in mixed-curvature spaces to capture complex structural patterns and combines modality- and space-level fusion.

Several fundamental limitations persist. Most methods are evaluated primarily on benchmarks where entities possess complete modalities, and relatively little attention has been given to scenarios with inherent modality asymmetry, such as fine art (e.g., visual artworks vs. text-only artists' names) or medicine (e.g., image-based diagnoses vs. text-only symptoms). Moreover, while approaches such as LAFA~\cite{lafa}, MCKGC~\cite{mckgc}, TCL~\cite{tcl}, and MR-MKG~\cite{mr_mkg} attempt to capture within-modal and cross-modal interactions or improve alignment, they typically focus on knowledge graph completion tasks under the assumption of complete modalities, and have not been extensively evaluated on knowledge graphs with systematic modality asymmetry. Meanwhile, works like GraphCLIP~\cite{graphclip}, VISTA~\cite{vista}, and MarT~\cite{mart} leverage vision-language pretraining, but they are limited in various ways: GraphCLIP lacks explicit relational modeling, VISTA lacks heterogeneous entity support, and MarT does not address modality asymmetry. Our work addresses this gap through VL-KGE, which integrates vision-language pretraining into a relationally grounded KGE framework, bridging structural modeling and cross-modal alignment.

\subsection{Vision--Language Models (VLMs)}\label{subsec:vlms}

Vision-language models (VLMs) have emerged as a powerful paradigm for multimodal representation learning, aligning visual and textual information within shared embedding spaces. Models such as CLIP~\cite{clip} and ALIGN~\cite{align} demonstrate the effectiveness of large-scale contrastive pretraining, while later approaches including LiT~\cite{lit}, SimVLM~\cite{simvlm}, BLIP~\cite{blip}, and BLIP-2~\cite{blip_2} improve transferability through generative objectives. Beyond contrastive learning, fusion-based architectures such as ViLBERT~\cite{vilbert}, FLAVA~\cite{flava}, and UniT~\cite{unit} capture fine-grained cross-modal interactions through co-attention or dual-stream transformers, while others extend language backbones for image captioning and visual question answering~\cite{visual_gpt, frozen}. Despite their success, most VLMs are optimized for open-ended tasks~\cite{vqa, vqa_2, vqa_gnn}, and they are not designed to model relational structure or support heterogeneous entity types. Recent methods, such as GraphCLIP~\cite{graphclip}, Knowledge-CLIP~\cite{knowledge_clip}, and MKGRL-MS~\cite{mkgrl_ms}, integrate VLMs with graph structure, but assume complete modalities across entities. In contrast, we leverage pretrained VLMs within a structured KGE framework to handle modality asymmetry and enable unified cross-modal representations.

\subsection{Computational Fine Art Analysis}\label{subsec:fine_art_analysis_kg}

Computational fine art analysis increasingly leverages visual content and structured knowledge. Methods relying primarily on visual content~\cite{tan_artgan, wilber_bam, elgammal2018shape, deepart, strezoski_omniart, artemis_v2, set2seq_transformer} achieve success in diverse tasks, but do not explicitly model the semantic relationships that contextualize artworks. Multimodal methods like ArtGraph~\cite{artgraph}, GCNBoost~\cite{gcnboost}, and ArtSAGENet~\cite{artsagenet} combine graph structure with visual representations, while GraphCLIP~\cite{graphclip} and knowledge-grounded description methods~\cite{explain_me_the_painting, artrag} leverage vision-language pretraining to improve artwork classification and understanding. We instead focus on learning fine art knowledge graph embeddings under modality asymmetry. Unlike standard KGE benchmarks~\cite{wn9_img_dataset, tbkge, mmkg}, where entities possess complete modalities, fine art knowledge graphs naturally exhibit modality asymmetry across entity types. To facilitate research on multimodal KGE under inherent modality asymmetry, we introduce two large-scale fine art knowledge graphs, WikiArt-MKG-v1 and WikiArt-MKG-v2.

\section{VL-KGE}\label{sec:approach}

We propose VL-KGE, a multimodal knowledge graph embedding method that integrates pretrained vision-language representations with structured relational modeling. VL-KGE is designed to address modality asymmetry in real-world knowledge graphs, where entities contain different modalities. By integrating pretrained vision-language representations with structural embeddings, VL-KGE learns unified multimodal representations that preserve relational structure and support inductive inference over unseen entities.

\subsection{Overview and Problem Formulation}\label{subsec:overview_problem_formulation}

\textbf{Problem formulation.} Let $\mathcal{G} = (\mathcal{E}, \mathcal{R}, \mathcal{T})$ denote a knowledge graph, where $\mathcal{E}$ is the set of entities, $\mathcal{R}$ the set of relations, and $\mathcal{T} \subseteq \mathcal{E} \times \mathcal{R} \times \mathcal{E}$ the set of triples. Each triple $(h, r, t) \in \mathcal{T}$ connects a head entity $h \in \mathcal{E}$ to a tail entity $t \in \mathcal{E}$ through a relation $r \in \mathcal{R}$. Entities may be associated with an image $I_e$, a textual description $T_e$, both, or neither. Our objective is to learn a scoring function $f(h, r, t)$ that assigns high scores to positive triples and low scores to negative ones. The key challenge is to effectively represent each entity $e$ as a single point in a unified embedding space by fusing its available modalities, i.e., structural, visual, and textual, such that all available information about the entity $e$ is integrated into the same representation. This unified representation must handle varying modality coverage across entities while preserving the relational structure of the knowledge graph.

\textbf{Multimodal encoders.} Each entity $e \in \mathcal{E}$ may be associated with one or more modalities. We denote by $\mathbf{s}_e \in \mathbb{R}^d$ a trainable structural embedding, by $\mathbf{v}_e = f_{\mathrm{img}}(I_e) \in \mathbb{R}^{d}$ a visual embedding obtained from a pretrained image encoder, and by $\mathbf{t}_e = f_{\mathrm{text}}(T_e) \in \mathbb{R}^{d}$ a textual embedding produced by a pretrained text encoder. Let $\mathcal{M}_e \subseteq \{\mathrm{structural}, \mathrm{visual}, \mathrm{textual}\}$ denote the set of available modalities for entity $e$. The pretrained encoders are either kept frozen to retain broad cross-modal alignment and enable inductive inference, or fine-tuned jointly by optimizing the downstream objective to improve domain adaptation. When pretrained encoders produce embeddings of different dimensionalities, lightweight linear projections map them into a shared $d$-dimensional space.

\textbf{Relational modeling.} The unified entity embeddings are integrated with standard KGE backbones to capture relational semantics. Relation embeddings $\mathbf{r} \in \mathbb{R}^d$ are treated as trainable parameters, learned jointly with the rest of the model. VL-KGE is compatible with widely used methods such as TransE~\cite{transe}, DistMult~\cite{distmult}, ComplEx~\cite{complex}, and RotatE~\cite{rotate} without modifying their scoring functions. Our framework replaces traditional entity representations with multimodal ones, enabling cross-modal modeling while preserving the relational inductive biases of each backbone.

\subsection{Inductive Entity Representation} \label{subsec:inductive_entity_representation}
A key feature of VL-KGE is its ability to perform inductive inference, making predictions for entities unseen during training. Unlike traditional KGE methods, which require retraining when new entities are added, VL-KGE directly leverages pretrained vision-language representations for new entities without introducing additional entity-specific parameters.

\textbf{Entity representation.} Let $\delta_e \in \{0,1\}$ indicate whether entity $e$ was observed during training. Structural embeddings are masked for unseen entities:
\begin{equation}
\hat{\mathbf{s}}_e = \delta_e \mathbf{s}_e,
\end{equation}
and combined with available modality features:
\begin{equation} \label{eq:entity_representation}
\mathbf{x}_e^{(m)} =
\begin{cases}
\hat{\mathbf{s}}_e, & \text{if } m = \mathrm{structural},\\[4pt]
\mathbf{v}_e, & \text{if } m = \mathrm{visual},\\[4pt]
\mathbf{t}_e, & \text{if } m = \mathrm{textual},
\end{cases}
\quad \text{for } m \in \mathcal{M}_e.
\end{equation}
These are fused to obtain the entity representation $\boldsymbol{r}_e$ as follows:
\begin{equation} \label{eq:fusion}
\boldsymbol{r}_e = \mathcal{F}\bigl(\{\mathbf{x}_e^{(m)} \mid m \in \mathcal{M}_e\}\bigr),
\end{equation}
where $\mathcal{F}$ is a fusion operator detailed in Section~\ref{subsec:fusion_mechanisms}. For unseen entities ($\delta_e=0$), the representation depends entirely on pretrained features, enabling inductive inference.

\textbf{Complex-valued backbones (ComplEx, RotatE).} For models requiring complex embeddings $\mathbf{z}_e = \boldsymbol{r}_e + i\mathbf{u}_e$, we construct the imaginary component $\mathbf{u}_e$ to preserve inductiveness. For entities with a learned imaginary component, we retrieve it from a parameterized embedding matrix; otherwise, we derive it from the real component via a shared projection:
\begin{equation}
\mathbf{u}_e =
\begin{cases}
\mathbf{l}_e, & \text{if } e \text{ has a learned imaginary component},\\[6pt]
\gamma P\boldsymbol{r}_e, & \text{otherwise},
\end{cases}
\end{equation}
where $\mathbf{l}_e$ is the learned imaginary embedding for entity $e$, $P \in \mathbb{R}^{d \times d}$ is a shared real-to-imaginary projection matrix, $\gamma = \tanh \beta$ is a learned gating parameter, and $\beta$ is a single global scalar. For complex-valued backbones, the projection matrix $P$ is trained on observed entities and generalizes to unseen ones, preserving the expressive power of ComplEx and RotatE.

\textbf{Real-valued backbones (TransE, DistMult).} For real-valued backbones, we directly use $\boldsymbol{r}_e$ from Eq.~\eqref{eq:fusion} as the entity embedding.

\subsection{Cross-Modal Fusion Mechanisms}\label{subsec:fusion_mechanisms}

The fusion operator $\mathcal{F}$ in Eq.~\eqref{eq:fusion} combines all available modalities into a unified entity representation, enabling consistent integration of VLM-derived representations with relational structure while supporting entities with missing modalities. Structural embeddings are learned jointly with relation embeddings, while visual and textual embeddings are extracted from pretrained VLM encoders that are either kept frozen or optionally fine-tuned during training. When dimensionalities differ, lightweight linear projections are used to map modality-specific features into a unified $d$-dimensional space so that they can be fused and optimized jointly under the KGE objective. We consider three fusion strategies. Average fusion computes the mean across all available modalities:
\begin{equation}
\boldsymbol{r}_e = \frac{1}{|\mathcal{M}_e|} \sum_{m \in \mathcal{M}_e} \mathbf{x}_e^{(m)},
\end{equation}
where $\mathcal{M}_e$ is the set of modalities present for entity $e$. Concatenation fusion stacks available embeddings:
\begin{equation}
\boldsymbol{r}_e = \oplus_{m \in \mathcal{M}_e} \mathbf{x}_e^{(m)},
\end{equation}
where $\oplus$ denotes concatenation. Weighted fusion learns the relative importance of modalities:
\begin{equation}
\boldsymbol{r}_e = \sum_{m \in \mathcal{M}_e} \alpha_m \mathbf{x}_e^{(m)},
\end{equation}
where $\{\alpha_m\}_{m \in \mathcal{M}_e}$ are learnable weights capturing the relative importance of each modality. Average and weighted fusion handle missing modalities by operating only on available components, while concatenation uses zero-padding to maintain dimensional consistency. Results comparing different fusion strategies are shown in Table~\ref{tab:wn9_img_dataset_ablation} in the supplementary material.

\subsection{Training Objective}\label{subsec:training_objective}
We train VL-KGE to assign higher scores to positive triples than to negative ones using the logistic loss. Let $\Omega^+$ denote the set of observed triples and $\Omega^-$ the set of negatives generated by randomly corrupting either the head or the tail entity. The loss is defined as:
\begin{equation}
\mathcal{L} = \sum_{(h, r, t) \in \Omega^+ \cup \Omega^-} \log \left( 1 + \exp(-y \cdot f(h, r, t)) \right),
\end{equation}
where $y \in \{+1, -1\}$ denotes the label for positive and negative triples, and $f(h, r, t)$ is the backbone-specific scoring function.

\section{Experimental Setup}\label{sec:experimental_setup}

In this section, we describe the datasets, baselines, evaluation protocol, and implementation details used in our experiments.

\subsection{Datasets}\label{subsec:datasets}

\textbf{WN9-IMG}~\cite{wn9_img_dataset} is a multimodal KG derived from WordNet~\cite{wordnet}, where entities correspond to ImageNet~\cite{imagenet} synsets. Each entity possesses both visual representations and textual descriptions, connected through semantic relations such as hypernymy (\textit{is-a}) and meronymy (\textit{has-part}). WN9-IMG provides a controlled setting for evaluating multimodal integration when modality coverage is complete and uniform across all entities (Figure~\ref{fig:wn9_img_dataset_task}).

\textbf{WikiArt-MKGs.} We introduce two fine art multimodal KGs that reframe traditional classification tasks as link prediction problems. Unlike WN9-IMG, where all entities have both visual and textual representations, artworks in WikiArt-MKGs are predominantly visual, while artists, movements, and other attributes are primarily textual, with many entities lacking one or more modalities entirely. This heterogeneity and modality asymmetry make WikiArt-MKG-v1 and WikiArt-MKG-v2 substantially more challenging and closer to real-world multimodal KG scenarios (Figure~\ref{fig:wikiart_mkg_v1_v2}). Statistics for all MKGs are provided in Table~\ref{tab:dataset_statistics}.

\begin{table}[t]
\centering
\caption{Statistics of MKGs. \textit{Visual} and \textit{Textual} denote the number of visual and textual entities, \textit{Entities} the total number of entities, \textit{Relations} the number of relation types, and \textit{Train}, \textit{Val}, and \textit{Test} the number of triples in each split.}
\label{tab:dataset_statistics}
\resizebox{\columnwidth}{!}{\begin{tabular}{@{} lccccccc @{}}
\toprule
MKG & Visual & Textual & Entities & Relations & Train & Val & Test \\ \midrule
WN9-IMG          & 6,555   & 6,555 & 6,555       & 9              & 11,741           & 1,337   & 1,319   \\ 
WikiArt-MKG-v1   & 75,921 & 837   & 76,758      & 4              & 299,968           & 34,020            & 19,695   \\ 
WikiArt-MKG-v2    & 216,564 & 7,602  & 224,166     & 22              & 7,877,220           & 208,513           & 208,368   \\ 
\bottomrule
\end{tabular}}
\end{table}

\textbf{WikiArt-MKG-v1} is a multimodal KG that models artworks (visual entities) as head entities and their attributes, i.e., artists, styles, creation years, and tags (textual entities), as tail entities, connected through task-based relations (\textit{isCreatedByArtist}, \textit{hasStyle}, \textit{isCreatedInYear}, \textit{isAssociatedWithTag}). This formulation transforms attribute prediction into structured link prediction. Modality availability is determined by entity type: artworks use visual representations, while all other entities (e.g., artists and styles) use textual representations. WikiArt-MKG-v1 comprises 76K artworks and 750 artists across four relation types.

\textbf{WikiArt-MKG-v2} builds on WikiArt-MKG-v1 and substantially extends it through large-scale web scraping from the WikiArt online collection~\cite{wikiart}, resulting in a large and diverse fine art KG with 217K artworks and 4K artists spanning 22 relation types. It introduces three key extensions: (i) enriched metadata including medium, location, nationality, dates, and institutional affiliations; (ii) diverse relational structure across artworks and artists, including artwork-to-attribute relations (e.g., \textit{isCreatedByArtist}, \textit{belongsToGenre}), artwork-to-artwork relations (\textit{isRelatedToArtwork}), and artist-to-artist relations (e.g., \textit{isInfluencedBy}, \textit{isPupilOf}, \textit{isRelatedToArtist}); and (iii) modality sparsity, where some entities lack visual or textual representations. The dataset is partitioned to enable multiple inductive evaluation settings. All test artworks are unseen during training. For artist-to-artist relations, we use disjoint artist subsets: triples connecting training and test artists are removed. For artwork-to-artist relations, artists appear in the training set through their links to seen artworks, but their links to test artworks are held out.

\subsection{Baselines}\label{subsec:baselines}

We instantiate VL-KGE with four standard KGE methods: TransE~\cite{transe}, DistMult~\cite{distmult}, ComplEx~\cite{complex}, and RotatE~\cite{rotate}, to disentangle the contribution of vision-language representations from that of relational modeling. For each backbone, we evaluate three encoder configurations: VB-KGE, VL-KGE with BLIP, and VL-KGE with CLIP. VB-KGE uses a ViT~\cite{vit} visual encoder pretrained on ImageNet~\cite{imagenet} and a BERT~\cite{bert} textual encoder, which are used independently without cross-modal alignment. VL-KGE uses either BLIP~\cite{blip}, a generative vision-language model, or CLIP~\cite{clip}, a contrastive vision-language model, as base encoders. For WN9-IMG, we additionally compare against MMKRL~\cite{mmkrl} and OTKGE~\cite{otkge}, two recent multimodal KGE methods designed for complete modality coverage. For WikiArt-MKGs, we also evaluate zero-shot baselines by computing cosine similarity between visual representations of artworks and textual prompts (e.g., ``This painting was created by artist X.'') for artwork-to-attribute prediction, and between textual representations of artists and textual prompts (e.g., ``This artist is influenced by artist X.'') for artist-to-artist prediction. Linear-probe results for visual encoders on the WikiArt-v1 and WikiArt-v2 datasets are provided in Table~\ref{tab:linear_probe_results} in the supplementary material.

\subsection{Evaluation Protocol}\label{subsec:evaluation_protocol}

We follow the standard filtered link prediction setting~\cite{otkge}, removing all true triples from the candidate set except the query triple, and report mean reciprocal rank (MRR) and Hits@$K$ for $K \in \{1, 3, 10\}$. For WN9-IMG, we adopt the standard bidirectional protocol~\cite{transe}, ranking candidates for both head $(?, r, t)$ and tail $(h, r, ?)$ prediction. For WikiArt-MKGs, we focus on tail prediction $(h, r, ?)$, as this reflects the primary inference tasks, i.e., predicting an artwork's artist, creation period, or stylistic attributes, with candidates selected per relation type. For WikiArt-MKG-v2, all test triples are inductive by construction. Across relation types, we distinguish between cases where participating entities are partially observed during training (e.g., artwork-to-artist) and cases where they are entirely unseen (e.g., artwork-to-artwork, artist-to-artist). Two relations, \textit{isRelatedToArtwork} and \textit{isRelatedToArtist}, are evaluated as semantic retrieval tasks, and the results are reported in Tables~\ref{tab:wikiart_v2_related_artworks} and~\ref{tab:wikiart_v2_related_artists} in the supplementary material.

\subsection{Implementation Details}\label{subsec:implementation_details}

All experiments are implemented in PyTorch~\cite{pytorch} and conducted on a single NVIDIA A100 GPU. Structural entity and relation embeddings are randomly initialized and optimized using Adagrad~\cite{adagrad} with a learning rate of $\eta = 0.1$ and a batch size of 512 triples. While prior work has shown that KGE performance is highly sensitive to hyperparameter choices~\cite{understanding_the_performance_kge, training_kges}, tuning them individually often benefits specific relations at the expense of others, particularly in our inductive, modality-asymmetric setup. Since our aim is to assess the relative gains provided by VLM-based representations rather than to maximize performance of individual backbones, we fix all training hyperparameters globally across all methods. Training efficiency scales linearly with the dataset size. On WN9-IMG, training takes approximately 20 seconds per epoch with 1 GB peak GPU memory and 5M trainable parameters. On WikiArt-MKG-v2, training takes approximately 3.5 minutes per epoch with 4.5 GB peak memory and 172M trainable parameters.

\section{Results}\label{sec:results}

\begin{table}[t]
\centering
\caption{Results on WN9-IMG. MMKRL and OTKGE results are as reported in~\cite{otkge}. For all metrics, higher is better. For each KGE, best results are in bold, second best \underline{underlined}.}
\label{tab:wn9_img_dataset_results}
\resizebox{\columnwidth}{!}{%
\begin{tabular}{@{} lcccc @{}}
\toprule
Method & MRR & Hits@1 & Hits@3 & Hits@10 \\
\midrule
MMKRL & 0.913 & 0.905 & 0.917 & 0.932 \\
OTKGE & 0.923 & 0.911 & 0.930 & 0.947 \\
\midrule
TransE & 0.904 & \textbf{0.894} & 0.909 & 0.922 \\
VB-TransE & \underline{0.910} & \underline{0.890} & \underline{0.923} & \underline{0.944} \\
VL-TransE (base: BLIP) & \underline{0.910} & \textbf{0.894} & 0.921 & 0.940 \\
VL-TransE (base: CLIP) & \textbf{0.913} & \underline{0.890} & \textbf{0.928} & \textbf{0.950} \\
\midrule
DistMult & 0.904 & 0.902 & 0.904 & 0.907 \\
VB-DistMult & \underline{0.923} & \underline{0.914} & \underline{0.927} & \underline{0.938} \\
VL-DistMult (base: BLIP) & 0.909 & 0.907 & 0.908 & 0.914 \\
VL-DistMult (base: CLIP) & \textbf{0.935} & \textbf{0.925} & \textbf{0.940} & \textbf{0.957} \\
\midrule
ComplEx & 0.900 & 0.899 & 0.901 & 0.902 \\
VB-ComplEx & \underline{0.916} & \underline{0.910} & \underline{0.918} & \underline{0.924} \\
VL-ComplEx (base: BLIP) & 0.903 & 0.900 & 0.904 & 0.907 \\
VL-ComplEx (base: CLIP) & \textbf{0.927} & \textbf{0.920} & \textbf{0.929} & \textbf{0.941} \\
\midrule
RotatE & 0.910 & \textbf{0.907} & 0.911 & 0.917 \\
VB-RotatE & 0.910 & 0.903 & \underline{0.914} & 0.925 \\
VL-RotatE (base: BLIP) & \underline{0.911} & 0.898 & \textbf{0.918} & \underline{0.931} \\
VL-RotatE (base: CLIP) & \textbf{0.914} & \underline{0.904} & \textbf{0.918} & \textbf{0.934} \\
\bottomrule
\end{tabular}}
\end{table}

\begin{table*}[t]
\centering
\caption{Results on WikiArt-MKGs. For all metrics, higher is better. For each KGE, best results are in bold, second best \underline{underlined}.}
\label{tab:wikiart_v1_v2_results}
\begin{tabular*}{\linewidth}{@{} l @{\extracolsep{\fill}} cccccccc @{}}
\toprule
Method & \multicolumn{4}{c}{WikiArt-MKG-v1} & \multicolumn{4}{c}{WikiArt-MKG-v2} \\
\cmidrule(r){2-5} \cmidrule(l){6-9}
   & MRR & Hits@1 & Hits@3 & Hits@10 & MRR & Hits@1 & Hits@3 & Hits@10 \\ \midrule
Zero-shot BLIP & 0.403 & 0.246 & 0.492 & 0.721 & 0.175 & 0.096 & 0.193 & 0.334 \\
Zero-shot CLIP & 0.510 & 0.357 & 0.609 & 0.798 & 0.237 & 0.139 & 0.263 & 0.442 \\
\midrule
TransE & 0.245 & 0.116 & 0.282 & 0.511 & 0.206 & 0.130 & 0.222 & 0.365 \\
VB-TransE & 0.489 & 0.325 & 0.584 & 0.833 & 0.309 & 0.186 & 0.356 & 0.558 \\
VL-TransE (base: BLIP) & \underline{0.681} & \textbf{0.537} & \underline{0.794} & \underline{0.924} & \underline{0.419} & \underline{0.307} & \underline{0.470} & \underline{0.641} \\
VL-TransE (base: CLIP) & \textbf{0.683} & \underline{0.535} & \textbf{0.799} & \textbf{0.938} & \textbf{0.526} & \textbf{0.399} & \textbf{0.597} & \textbf{0.772} \\
\midrule
DistMult & 0.135 & 0.044 & 0.118 & 0.342 & 0.157 & 0.095 & 0.151 & 0.305 \\
VB-DistMult & 0.649 & 0.503 & 0.756 & 0.911 & 0.461 & 0.346 & 0.518 & 0.689 \\
VL-DistMult (base: BLIP) & \underline{0.705} & \underline{0.565} & \underline{0.818} & \underline{0.940} & \underline{0.508} & \underline{0.392} & \underline{0.571} & \underline{0.731} \\
VL-DistMult (base: CLIP) & \textbf{0.781} & \textbf{0.659} & \textbf{0.888} & \textbf{0.974} & \textbf{0.577} & \textbf{0.462} & \textbf{0.643} & \textbf{0.796} \\
\midrule
ComplEx & 0.133 & 0.042 & 0.116 & 0.341 & 0.157 & 0.096 & 0.152 & 0.305 \\
VB-ComplEx & 0.649 & 0.504 & 0.758 & 0.912 & 0.463 & 0.348 & 0.520 & 0.692 \\
VL-ComplEx (base: BLIP) & \underline{0.715} & \underline{0.576} & \underline{0.830} & \underline{0.945} & \underline{0.515} & \underline{0.399} & \underline{0.579} & \underline{0.737} \\
VL-ComplEx (base: CLIP) & \textbf{0.785} & \textbf{0.665} & \textbf{0.889} & \textbf{0.975} & \textbf{0.578} & \textbf{0.465} & \textbf{0.642} & \textbf{0.795} \\
\midrule
RotatE & 0.276 & 0.136 & 0.327 & 0.563 & 0.215 & 0.135 & 0.232 & 0.387 \\
VB-RotatE & 0.491 & 0.328 & 0.586 & 0.834 & 0.414 & 0.300 & 0.463 & 0.649 \\
VL-RotatE (base: BLIP) & \underline{0.670} & \underline{0.520} & \underline{0.788} & \underline{0.925} & \textbf{0.459} & \textbf{0.341} & \textbf{0.518} & \textbf{0.690} \\
VL-RotatE (base: CLIP) & \textbf{0.724} & \textbf{0.590} & \textbf{0.833} & \textbf{0.950} & \underline{0.439} & \underline{0.326} & \underline{0.489} & \underline{0.667} \\
\bottomrule
\end{tabular*}
\end{table*}

\begin{figure*}[t]
\centering
% Preamble:
% \usepackage{graphicx,booktabs,multirow,tabularx,array}
% (no floats here; this works inside adjustbox/minipage/columns)

\newlength{\thumbgap}\setlength{\thumbgap}{2pt}
\newcommand{\fiveThumbs}[5]{%
  \begingroup
  \setlength{\fboxsep}{0pt}%
  \def\thumbw{\dimexpr(\linewidth-4\thumbgap)/5\relax}%
  \includegraphics[width=\thumbw]{#1}\hspace{\thumbgap}%
  \includegraphics[width=\thumbw]{#2}\hspace{\thumbgap}%
  \includegraphics[width=\thumbw]{#3}\hspace{\thumbgap}%
  \includegraphics[width=\thumbw]{#4}\hspace{\thumbgap}%
  \includegraphics[width=\thumbw]{#5}%
  \endgroup
}

% --- NEW: smaller entity value text ---
\newcommand{\valsize}{\scriptsize} % change to \scriptsize for tighter fit
\newcommand{\vals}[1]{%
  \par\noindent\valsize\RaggedRight #1\par
}

% One relation cell: centered label, left-aligned values, then a thin rule
\newcommand{\relcell}[2]{%
  \parbox{\linewidth}{%
    {\centering \footnotesize #1\par}% end centered paragraph
    \vals{#2}% start values as their own paragraph
    \vspace{-2pt}%
    \noindent\rule{\linewidth}{0.4pt}\par\vspace{2pt}%
  }%
}

% Same, but for the thumbnails row
\newcommand{\relcellthumbs}[2]{%
  \parbox{\linewidth}{%
    {\centering \footnotesize #1\par}%
    \vspace{1pt}
    \par\noindent #2\par%
  }%
}

% Without separator line (use only for the final row before \bottomrule)
\newcommand{\relcellend}[2]{%
  \parbox{\linewidth}{%
    {\centering \footnotesize #1\par}%
    \vals{#2}%
  }%
}

\setlength{\tabcolsep}{6pt}
\renewcommand\arraystretch{1.3}
\begin{tabularx}{\textwidth}{>{\centering\arraybackslash}X c >{\centering\arraybackslash}X}
  \toprule
 Zero-shot CLIP & Query & VL-ComplEx \\
  \midrule

  %%%%%%%%%%%%%%%%%%%%%%%%%%%%%%%
  % Center column spans ALL rows
  %%%%%%%%%%%%%%%%%%%%%%%%%%%%%%%
  % 10 rows total: 5 (artwork) + 5 (artist)

  % --- isCreatedByArtist (artwork) ---
  \relcell{isCreatedByArtist}{berthold-woltze; marten-van-mytens-the-younger; cornelis-cort; theodoor-van-thulden; joachim-govertsz-camphuysen}
  &
  \multirow{10}{*}{%
    \begin{minipage}{0.2\textwidth}\centering
      \includegraphics[width=\linewidth]{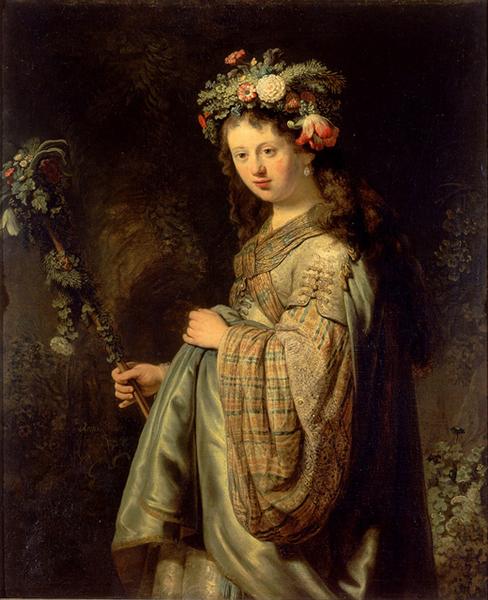}\\[2pt]
      \small \emph{Flora} (1634)\\[40pt]
      \includegraphics[width=.7\linewidth, height=2.5cm]{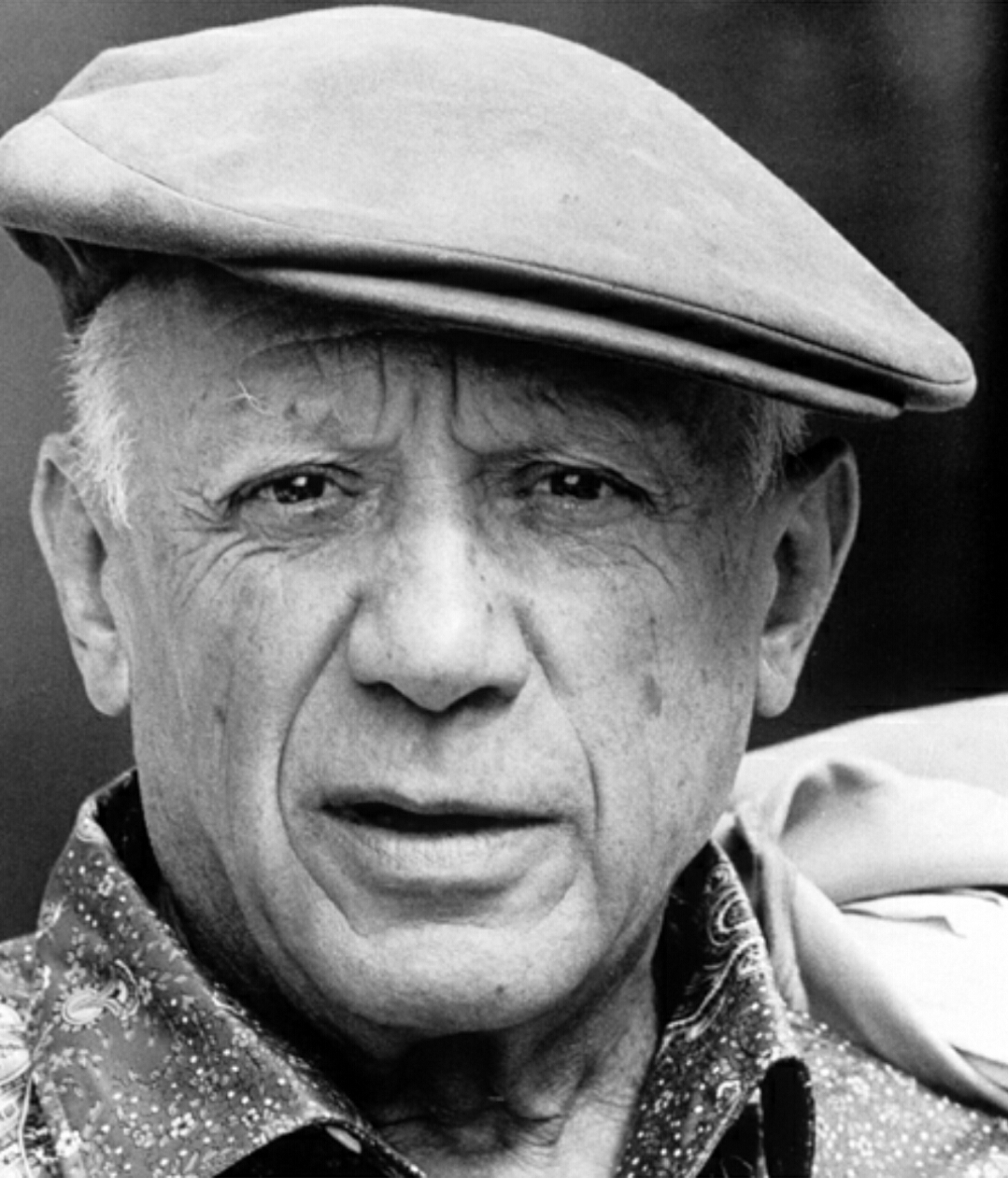}\\[2pt]
      \small Pablo Picasso
    \end{minipage}
  }
  &
  \relcell{isCreatedByArtist}{\textbf{rembrandt}; frans-van-mieris-the-elder; gerrit-dou; ferdinand-bol; pieter-soutman}
  \\[2pt]

  % --- hasStyle (artwork) ---
  \relcell{hasStyle}{\resizebox{\linewidth}{!}{northern-renaissance; \textbf{baroque}; rococo; high-renaissance; early-renaissance}}
  &&
  \relcell{hasStyle}{\textbf{baroque}; romanticism; realism; fauvism; rococo}
  \\[2pt]

  % --- isCreatedInYear (artwork) ---
  \relcell{isCreatedInYear}{1626; 1625; 1651; 1641; 1627}
  &&
  \relcell{isCreatedInYear}{1640; 1633; 1631; 1657; 1630}
  \\[2pt]

  % --- isAssociatedWithTag (artwork) ---
  \relcell{isAssociatedWithTag}{st.-barbara; esther; pygmalion; j.walch. "de-vreeselijke-avonturen-van-scholastica"; flower-arranging}
  &&
  \relcell{isAssociatedWithTag}{\textbf{lady}; female-portraits; plant; mythology; headpiece\vspace{\baselineskip}}
  \\[2pt]

  % --- Location Prediction (artwork) ---
  \relcell{isLocatedIn}{mauritshuis, hague, netherlands; latvian national museum of art, riga, latvia; centraal museum, utrecht, netherlands; gemeentemuseum den haag, hague, netherlands; rijksmuseum, amsterdam, netherlands}
  &&
  \relcell{isLocatedIn}{\textbf{hermitage museum, saint petersburg, russia}; private collection; gemäldegalerie, berlin, germany; national museum, warsaw, poland; mauritshuis, hague, netherlands}
  \\[2pt]

  % --- isRelatedToArtwork (artwork) ---
  \relcellthumbs{isRelatedToArtwork}{%
    \fiveThumbs{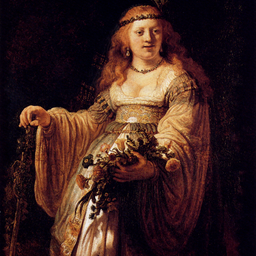}
               {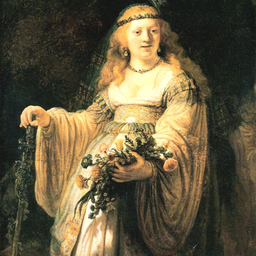}
               {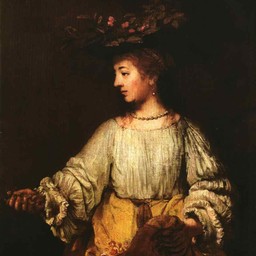}
               {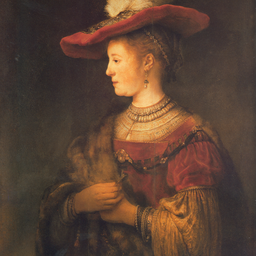}
               {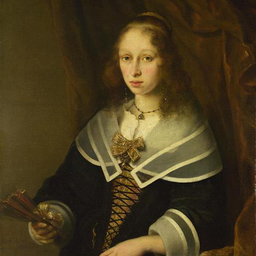}}
  &&
  \relcellthumbs{isRelatedToArtwork}{%
    \fiveThumbs{Images/rembrandt_saskia-van-uylenburgh-in-arcadian-costume-1635.png}
               {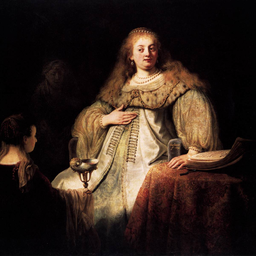}
               {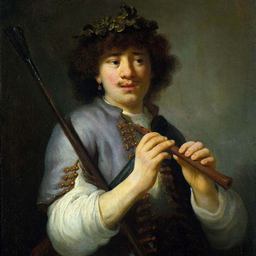}
               {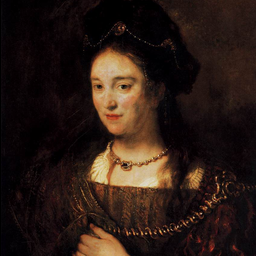}
               {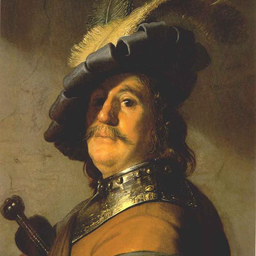}}
  \\

  %%%%%%%%%%%%%%%%%%%%%%%%%%%%%%%
  % Artist relations (no scores)
  %%%%%%%%%%%%%%%%%%%%%%%%%%%%%%%
  \midrule

  % --- isAssociatedWithArtMovement (artist) ---
  \relcell{isAssociatedWithArtMovement}{\textbf{cubism}; dada; new-european-painting; avant-garde; modernismo}
  &&
  \relcell{isAssociatedWithArtMovement}{\textbf{cubism}; romanticism; impressionism; \textbf{surrealism}; \textbf{post-impressionism}}
  \\[2pt]

  % --- isMemberOfPaintingSchool (artist) ---
  \relcell{isMemberOfPaintingSchool}{bauhaus; les-xx; degenerate-art; zero; cobra}
  &&
  \relcell{isMemberOfPaintingSchool}{\textbf{ecole-de-paris}; macchiaioli; degenerate-art; les-fauves; barbizon-school}
  \\[2pt]

  % --- isInfluencedBy (artist) ---
  \relcell{isInfluencedBy}{pablo-picasso; vincent-van-gogh; leonardo-da-vinci; charles-gleyre; xavier-mellery}
  &&
  \relcell{isInfluencedBy}{adam-elsheimer; camille-corot; nicholas-roerich; david-alfaro-siqueiros; henri-edmond-cross}
  \\[2pt]

  % --- isInfluencedOn (artist) ---
  \relcell{isInfluencedOn}{pablo-picasso; vincent-van-gogh; lo-scheggia; leonardo-da-vinci; john-quidor}
  &&
  \relcell{isInfluencedOn}{mikhail-larionov; \textbf{david-alfaro-siqueiros}; lucian-freud; ford-madox-brown; njideka-akunyili-crosby}
  \\[2pt]

  % --- isRelatedToArtist (artist) ---
  \relcellend{isRelatedToArtist}{pablo-picasso; alice-aycock; dominique-gonzalez-foerster; \textbf{jean-auguste-dominique-ingres}; monique-orsini}
  &&
  \relcellend{isRelatedToArtist}{\textbf{georges-braque}; ohara-koson; lyonel-feininger; jean-arp; \textbf{david-alfaro-siqueiros}}
  \\

  \bottomrule
\end{tabularx}

\caption{Qualitative comparison of zero-shot CLIP and VL-ComplEx (base: CLIP) on WikiArt-MKG-v2. Given an artwork (top rows) or an artist (bottom rows) as a query, we show the top-5 predicted entities for selected relations. For artist queries, we use only textual input representations. Correctly retrieved entities are shown in bold.}
\label{fig:qualitative_analysis_clip_complex}
\end{figure*}

We now present the experimental evaluation of VL-KGE across three multimodal KGs, examining its ability to integrate pretrained vision-language representations with relational structure under diverse modality configurations. We report quantitative results in Tables~\ref{tab:wn9_img_dataset_results} and~\ref{tab:wikiart_v1_v2_results} and qualitative analysis in Figure~\ref{fig:qualitative_analysis_clip_complex}, focusing on two key questions: (i) how pretrained vision-language encoders affect KGE performance across heterogeneous settings, and (ii) how VL-KGE performs under modality asymmetry.

\subsection{Results on WN9-IMG}\label{subsec:results_wn9_img}

\textbf{Performance under modality completeness.} Table~\ref{tab:wn9_img_dataset_results} reports results on WN9-IMG, a benchmark where all entities are associated with both visual and textual modalities. Across all four scoring functions, VL-KGE outperforms unimodal baselines, demonstrating the benefit of integrating pretrained multimodal representations with structural modeling. The performance improvements are especially notable for DistMult and ComplEx, where multimodal integration yields consistent gains, highlighting the synergy between pretrained representations and structural modeling. We also include MMKRL and OTKGE as representative multimodal KGE baselines on WN9-IMG; their reported results are slightly lower than those of the best VL-KGE variants, confirming that VL-KGE performs competitively compared to prior multimodal methods.

\textbf{Impact of pretraining alignment.} A key observation is that CLIP-based VL-DistMult achieves the strongest overall performance, illustrating how pretrained vision-language representations enhance relational reasoning. Interestingly, models based on the decoupled ViT+BERT configuration often perform competitively with or occasionally surpass BLIP-based variants. This behavior is likely attributable to the nature of WN9-IMG: since the dataset is derived from ImageNet synsets, ViT's ImageNet pretraining provides strong domain alignment, yielding highly discriminative visual representations. This result underscores the importance of domain alignment between the pretrained encoders and the downstream knowledge graph, suggesting that such alignment has a considerable impact on relational reasoning. Overall, these findings demonstrate that VL-KGE effectively leverages multimodal signals when all modalities are available for all entities, while also highlighting the significance of backbone selection in maximizing downstream performance.

\subsection{Results on WikiArt-MKGs}\label{subsec:results_wikiart_mkgs}

\textbf{Zero-shot baseline performance.} Table~\ref{tab:wikiart_v1_v2_results} reports results on WikiArt-MKG-v1 and WikiArt-MKG-v2. We begin with zero-shot baselines that use frozen vision-language representations without any KGE training. Despite their simplicity, these approaches achieve non-trivial link prediction performance, showing that pretrained VLMs encode semantic structure relevant to relational reasoning. CLIP consistently outperforms BLIP across both datasets, reflecting stronger alignment between visual and textual modalities. However, absolute performance remains limited, demonstrating that pretrained representations alone are insufficient to fully capture the relational structure of the underlying knowledge graphs.

\textbf{Impact of multimodal integration.} Integrating pretrained multimodal representations with structural modeling yields substantial performance improvements over unimodal baselines. On WikiArt-MKG-v1, which contains artwork-to-attribute relations where artworks are visual and attributes are textual, VL-KGE consistently surpasses both structural-only and multimodal baselines using separately pretrained encoders across all scoring functions, highlighting how aligned vision-language representations enable effective learning under modality asymmetry. The gains are especially pronounced for VL-DistMult and VL-ComplEx, which benefit significantly from the integration of vision-language representations. The advantages of multimodal integration become even more evident in WikiArt-MKG-v2, which features a larger number of entities, more complex relational structure, and extensive modality asymmetry across entities. In this challenging setting, VL-KGE not only handles missing modalities but leverages cross-modal signals to improve link prediction in large-scale complex knowledge graphs. These results demonstrate the robustness of VL-KGE in modality-asymmetric settings with heterogeneous entity types.

\textbf{Backbone considerations and model behavior.} CLIP-based variants of VL-KGE generally outperform those based on BLIP, reflecting CLIP's stronger vision-language alignment and broader domain coverage. Notably, CLIP-based VL-ComplEx achieves the best performance on WikiArt-MKG-v1, while CLIP-based VL-ComplEx and VL-DistMult perform comparably on WikiArt-MKG-v2. However, both CLIP- and BLIP-based VL-KGEs consistently outperform their VB-KGE counterparts that use separately pretrained encoders, confirming the importance of aligned vision-language representations for multimodal KGE. An interesting exception is VL-RotatE, where BLIP shows competitive performance with CLIP on WikiArt-MKG-v2, potentially due to the dataset's compositional artist-to-artist relations. Additional results in the supplementary material include fine-tuning ablation studies (Table~\ref{tab:wikiart_v1_v2_ablation}) demonstrating the effectiveness of pretrained vision-language representations, and per-relation analysis (Figure~\ref{fig:mrr_per_relation}) showing that bilinear models excel at many-to-one relations while translational models perform better on sparse artist-to-artist relations.

\subsection{Qualitative Analysis}\label{subsec:qualitative_analysis}

To complement the quantitative evaluation, we perform a qualitative analysis to assess the nature and depth of relational knowledge captured by VL-KGE. Figure~\ref{fig:qualitative_analysis_clip_complex} presents representative outputs for both artwork- and artist-centric queries on the WikiArt-MKG-v2, contrasting zero-shot CLIP with CLIP-based VL-ComplEx. For each query entity, we examine predictions across multiple relations. The top rows show an artwork query, Rembrandt's \emph{Flora} (1634), while the bottom rows show an artist query, Pablo Picasso. For each relation, we display the top-5 predicted entities from both zero-shot CLIP and CLIP-based VL-ComplEx. Although zero-shot CLIP retrieves entities that are often visually plausible, its representations are largely restricted to low-level similarity cues and lack sensitivity to the structured dependencies inherent in the knowledge graph. This limitation is particularly evident in relation-centric predictions, where CLIP frequently produces self-referential or semantically inconsistent outputs (e.g., predicting the head entity as its own influence) and fails to recover historically or contextually appropriate entities. In contrast, VL-KGE leverages relational structure during training to align multimodal representations with the semantics of the underlying knowledge graph. This enables the model to recover entities that reflect not only visual or textual similarity but also historically grounded and semantically coherent relationships. For the artwork \emph{Flora} (1634), VL-KGE correctly associates the artwork with its creator (Rembrandt), situates it within the Baroque stylistic context, infers plausible temporal and geographic attributes, and retrieves thematically related artworks. For the artist query, VL-KGE captures higher-order dependencies such as movement affiliation (e.g., Cubism), membership in relevant art schools (e.g., École de Paris), connections to contemporaries (e.g., Georges Braque), and both antecedent and subsequent influences (e.g., Camille Corot and Lucian Freud). These results indicate that integrating pretrained vision-language representations with structural modeling improves encoding of domain-specific semantics. By explicitly incorporating graph structure, VL-KGE exhibits a capacity for relational reasoning that zero-shot approaches lack, yielding predictions that are more contextually grounded, historically faithful, and aligned with the ontological structure of the knowledge graph.

\section{Conclusion}\label{sec:conclusion}

We present VL-KGE, a multimodal knowledge graph embedding framework that integrates pretrained vision-language representations with structural modeling to address the challenges of modality asymmetry and cross-modal alignment. By leveraging pre-aligned multimodal encoders and flexible fusion mechanisms, VL-KGE produces unified entity representations that handle modality variation, while preserving relational structure and enabling inductive inference over unseen entities. Experimental results on WN9-IMG and our newly introduced fine art knowledge graphs, WikiArt-MKG-v1 and WikiArt-MKG-v2, demonstrate the effectiveness of VL-KGE over unimodal and multimodal baselines. These findings highlight the potential of integrating vision-language models into knowledge graph embeddings and lay the groundwork for future research on multimodal knowledge representation and completion at scale.

\balance

\bibliographystyle{ACM-Reference-Format}
\bibliography{bibliography}


\begin{thebibliography}{90}

%%% the Web address.

\ifx \showCODEN    \undefined \def \showCODEN     #1{\unskip}     \fi
\ifx \showISBNx    \undefined \def \showISBNx     #1{\unskip}     \fi
\ifx \showISBNxiii \undefined \def \showISBNxiii  #1{\unskip}     \fi
\ifx \showISSN     \undefined \def \showISSN      #1{\unskip}     \fi
\ifx \showLCCN     \undefined \def \showLCCN      #1{\unskip}     \fi
\ifx \shownote     \undefined \def \shownote      #1{#1}          \fi
\ifx \showarticletitle \undefined \def \showarticletitle #1{#1}   \fi
\ifx \showURL      \undefined \def \showURL       {\relax}        \fi
% invisible to TeX
\providecommand\bibfield[2]{#2}
\providecommand\bibinfo[2]{#2}
\providecommand\natexlab[1]{#1}
\providecommand\showeprint[2][]{arXiv:#2}

\bibitem[Antol et~al\mbox{.}(2015)]%
        {vqa}
\bibfield{author}{\bibinfo{person}{Stanislaw Antol}, \bibinfo{person}{Aishwarya Agrawal}, \bibinfo{person}{Jiasen Lu}, \bibinfo{person}{Margaret Mitchell}, \bibinfo{person}{Dhruv Batra}, \bibinfo{person}{C.~Lawrence Zitnick}, {and} \bibinfo{person}{Devi Parikh}.} \bibinfo{year}{2015}\natexlab{}.
\newblock \showarticletitle{VQA: Visual Question Answering}. In \bibinfo{booktitle}{\emph{Proceedings of the IEEE International Conference on Computer Vision (ICCV)}}.
\newblock


\bibitem[Bai et~al\mbox{.}(2021)]%
        {explain_me_the_painting}
\bibfield{author}{\bibinfo{person}{Zechen Bai}, \bibinfo{person}{Yuta Nakashima}, {and} \bibinfo{person}{Noa Garcia}.} \bibinfo{year}{2021}\natexlab{}.
\newblock \showarticletitle{Explain Me the Painting: Multi-Topic Knowledgeable Art Description Generation}. In \bibinfo{booktitle}{\emph{2021 IEEE/CVF International Conference on Computer Vision (ICCV)}}. \bibinfo{pages}{5402--5412}.
\newblock
\href{https://doi.org/10.1109/ICCV48922.2021.00537}{doi:\nolinkurl{10.1109/ICCV48922.2021.00537}}


\bibitem[Bonner et~al\mbox{.}(2022)]%
        {understanding_the_performance_kge}
\bibfield{author}{\bibinfo{person}{Stephen Bonner}, \bibinfo{person}{Ian~P. Barrett}, \bibinfo{person}{Cheng Ye}, \bibinfo{person}{Rowan Swiers}, \bibinfo{person}{Ola Engkvist}, \bibinfo{person}{Charles~Tapley Hoyt}, {and} \bibinfo{person}{William~L. Hamilton}.} \bibinfo{year}{2022}\natexlab{}.
\newblock \showarticletitle{Understanding the performance of knowledge graph embeddings in drug discovery}.
\newblock \bibinfo{journal}{\emph{Artificial Intelligence in the Life Sciences}}  \bibinfo{volume}{2} (\bibinfo{year}{2022}), \bibinfo{pages}{100036}.
\newblock
\showISSN{2667-3185}
\href{https://doi.org/10.1016/j.ailsci.2022.100036}{doi:\nolinkurl{10.1016/j.ailsci.2022.100036}}


\bibitem[Bordes et~al\mbox{.}(2013)]%
        {transe}
\bibfield{author}{\bibinfo{person}{Antoine Bordes}, \bibinfo{person}{Nicolas Usunier}, \bibinfo{person}{Alberto Garcia-Duran}, \bibinfo{person}{Jason Weston}, {and} \bibinfo{person}{Oksana Yakhnenko}.} \bibinfo{year}{2013}\natexlab{}.
\newblock \showarticletitle{Translating Embeddings for Modeling Multi-relational Data}. In \bibinfo{booktitle}{\emph{Advances in Neural Information Processing Systems}}, \bibfield{editor}{\bibinfo{person}{C.J. Burges}, \bibinfo{person}{L.~Bottou}, \bibinfo{person}{M.~Welling}, \bibinfo{person}{Z.~Ghahramani}, {and} \bibinfo{person}{K.Q. Weinberger}} (Eds.), Vol.~\bibinfo{volume}{26}. \bibinfo{publisher}{Curran Associates, Inc.}
\newblock
\urldef\tempurl%
\url{https://proceedings.neurips.cc/paper\_files/paper/2013/file/1cecc7a77928ca8133fa24680a88d2f9-Paper.pdf}
\showURL{%
\tempurl}


\bibitem[Cao et~al\mbox{.}(2022)]%
        {otkge}
\bibfield{author}{\bibinfo{person}{Zongsheng Cao}, \bibinfo{person}{Qianqian Xu}, \bibinfo{person}{Zhiyong Yang}, \bibinfo{person}{Yuan He}, \bibinfo{person}{Xiaochun Cao}, {and} \bibinfo{person}{Qingming Huang}.} \bibinfo{year}{2022}\natexlab{}.
\newblock \showarticletitle{OTKGE: Multi-modal Knowledge Graph Embeddings via Optimal Transport}. In \bibinfo{booktitle}{\emph{Advances in Neural Information Processing Systems}}, \bibfield{editor}{\bibinfo{person}{S.~Koyejo}, \bibinfo{person}{S.~Mohamed}, \bibinfo{person}{A.~Agarwal}, \bibinfo{person}{D.~Belgrave}, \bibinfo{person}{K.~Cho}, {and} \bibinfo{person}{A.~Oh}} (Eds.), Vol.~\bibinfo{volume}{35}. \bibinfo{publisher}{Curran Associates, Inc.}, \bibinfo{pages}{39090--39102}.
\newblock
\urldef\tempurl%
\url{https://proceedings.neurips.cc/paper\_files/paper/2022/file/ffdb280e7c7b4c4af30e04daf5a84b98-Paper-Conference.pdf}
\showURL{%
\tempurl}


\bibitem[Castellano et~al\mbox{.}(2022)]%
        {artgraph}
\bibfield{author}{\bibinfo{person}{Giovanna Castellano}, \bibinfo{person}{Vincenzo Digeno}, \bibinfo{person}{Giovanni Sansaro}, {and} \bibinfo{person}{Gennaro Vessio}.} \bibinfo{year}{2022}\natexlab{}.
\newblock \showarticletitle{Leveraging Knowledge Graphs and Deep Learning for automatic art analysis}.
\newblock \bibinfo{journal}{\emph{Knowledge-Based Systems}}  \bibinfo{volume}{248} (\bibinfo{year}{2022}), \bibinfo{pages}{108859}.
\newblock
\showISSN{0950-7051}
\href{https://doi.org/10.1016/j.knosys.2022.108859}{doi:\nolinkurl{10.1016/j.knosys.2022.108859}}


\bibitem[Chen et~al\mbox{.}(2022)]%
        {visual_gpt}
\bibfield{author}{\bibinfo{person}{Jun Chen}, \bibinfo{person}{Han Guo}, \bibinfo{person}{Kai Yi}, \bibinfo{person}{Boyang Li}, {and} \bibinfo{person}{Mohamed Elhoseiny}.} \bibinfo{year}{2022}\natexlab{}.
\newblock \showarticletitle{VisualGPT: Data-efficient Adaptation of Pretrained Language Models for Image Captioning}. In \bibinfo{booktitle}{\emph{2022 IEEE/CVF Conference on Computer Vision and Pattern Recognition (CVPR)}}. \bibinfo{pages}{18009--18019}.
\newblock
\href{https://doi.org/10.1109/CVPR52688.2022.01750}{doi:\nolinkurl{10.1109/CVPR52688.2022.01750}}


\bibitem[Chen et~al\mbox{.}(2025)]%
        {snag}
\bibfield{author}{\bibinfo{person}{Zhuo Chen}, \bibinfo{person}{Yin Fang}, \bibinfo{person}{Yichi Zhang}, \bibinfo{person}{Lingbing Guo}, \bibinfo{person}{Jiaoyan Chen}, \bibinfo{person}{Jeff~Z. Pan}, \bibinfo{person}{Huajun Chen}, {and} \bibinfo{person}{Wen Zhang}.} \bibinfo{year}{2025}\natexlab{}.
\newblock \showarticletitle{Noise-powered Multi-modal Knowledge Graph Representation Framework}. In \bibinfo{booktitle}{\emph{Proceedings of the 31st International Conference on Computational Linguistics}}, \bibfield{editor}{\bibinfo{person}{Owen Rambow}, \bibinfo{person}{Leo Wanner}, \bibinfo{person}{Marianna Apidianaki}, \bibinfo{person}{Hend Al-Khalifa}, \bibinfo{person}{Barbara~Di Eugenio}, {and} \bibinfo{person}{Steven Schockaert}} (Eds.). \bibinfo{publisher}{Association for Computational Linguistics}, \bibinfo{address}{Abu Dhabi, UAE}, \bibinfo{pages}{141--155}.
\newblock
\urldef\tempurl%
\url{https://aclanthology.org/2025.coling-main.11/}
\showURL{%
\tempurl}


\bibitem[Deng et~al\mbox{.}(2009)]%
        {imagenet}
\bibfield{author}{\bibinfo{person}{Jia Deng}, \bibinfo{person}{Wei Dong}, \bibinfo{person}{Richard Socher}, \bibinfo{person}{Li-Jia Li}, \bibinfo{person}{Kai Li}, {and} \bibinfo{person}{Li Fei-Fei}.} \bibinfo{year}{2009}\natexlab{}.
\newblock \showarticletitle{ImageNet: A large-scale hierarchical image database}. In \bibinfo{booktitle}{\emph{2009 IEEE Conference on Computer Vision and Pattern Recognition}}. \bibinfo{pages}{248--255}.
\newblock
\href{https://doi.org/10.1109/CVPR.2009.5206848}{doi:\nolinkurl{10.1109/CVPR.2009.5206848}}


\bibitem[Dettmers et~al\mbox{.}(2018)]%
        {conve}
\bibfield{author}{\bibinfo{person}{Tim Dettmers}, \bibinfo{person}{Pasquale Minervini}, \bibinfo{person}{Pontus Stenetorp}, {and} \bibinfo{person}{Sebastian Riedel}.} \bibinfo{year}{2018}\natexlab{}.
\newblock \showarticletitle{Convolutional 2D Knowledge Graph Embeddings}.
\newblock \bibinfo{journal}{\emph{Proceedings of the AAAI Conference on Artificial Intelligence}} \bibinfo{volume}{32}, \bibinfo{number}{1} (\bibinfo{date}{Apr.} \bibinfo{year}{2018}).
\newblock
\href{https://doi.org/10.1609/aaai.v32i1.11573}{doi:\nolinkurl{10.1609/aaai.v32i1.11573}}


\bibitem[Devlin et~al\mbox{.}(2019)]%
        {bert}
\bibfield{author}{\bibinfo{person}{Jacob Devlin}, \bibinfo{person}{Ming-Wei Chang}, \bibinfo{person}{Kenton Lee}, {and} \bibinfo{person}{Kristina Toutanova}.} \bibinfo{year}{2019}\natexlab{}.
\newblock \showarticletitle{{BERT}: Pre-training of Deep Bidirectional Transformers for Language Understanding}. In \bibinfo{booktitle}{\emph{Proceedings of the 2019 Conference of the North {A}merican Chapter of the Association for Computational Linguistics: Human Language Technologies, Volume 1 (Long and Short Papers)}}, \bibfield{editor}{\bibinfo{person}{Jill Burstein}, \bibinfo{person}{Christy Doran}, {and} \bibinfo{person}{Thamar Solorio}} (Eds.). \bibinfo{publisher}{Association for Computational Linguistics}, \bibinfo{address}{Minneapolis, Minnesota}, \bibinfo{pages}{4171--4186}.
\newblock
\href{https://doi.org/10.18653/v1/N19-1423}{doi:\nolinkurl{10.18653/v1/N19-1423}}


\bibitem[Dosovitskiy et~al\mbox{.}(2021)]%
        {vit}
\bibfield{author}{\bibinfo{person}{Alexey Dosovitskiy}, \bibinfo{person}{Lucas Beyer}, \bibinfo{person}{Alexander Kolesnikov}, \bibinfo{person}{Dirk Weissenborn}, \bibinfo{person}{Xiaohua Zhai}, \bibinfo{person}{Thomas Unterthiner}, \bibinfo{person}{Mostafa Dehghani}, \bibinfo{person}{Matthias Minderer}, \bibinfo{person}{Georg Heigold}, \bibinfo{person}{Sylvain Gelly}, \bibinfo{person}{Jakob Uszkoreit}, {and} \bibinfo{person}{Neil Houlsby}.} \bibinfo{year}{2021}\natexlab{}.
\newblock \showarticletitle{An Image is Worth 16x16 Words: Transformers for Image Recognition at Scale}. In \bibinfo{booktitle}{\emph{International Conference on Learning Representations}}.
\newblock
\urldef\tempurl%
\url{https://openreview.net/forum?id=YicbFdNTTy}
\showURL{%
\tempurl}


\bibitem[Duchi et~al\mbox{.}(2011)]%
        {adagrad}
\bibfield{author}{\bibinfo{person}{John Duchi}, \bibinfo{person}{Elad Hazan}, {and} \bibinfo{person}{Yoram Singer}.} \bibinfo{year}{2011}\natexlab{}.
\newblock \showarticletitle{Adaptive Subgradient Methods for Online Learning and Stochastic Optimization}.
\newblock \bibinfo{journal}{\emph{Journal of Machine Learning Research}} \bibinfo{volume}{12}, \bibinfo{number}{61} (\bibinfo{year}{2011}), \bibinfo{pages}{2121--2159}.
\newblock
\urldef\tempurl%
\url{http://jmlr.org/papers/v12/duchi11a.html}
\showURL{%
\tempurl}


\bibitem[Efthymiou et~al\mbox{.}(2024)]%
        {set2seq_transformer}
\bibfield{author}{\bibinfo{person}{Athanasios Efthymiou}, \bibinfo{person}{Stevan Rudinac}, \bibinfo{person}{Monika Kackovic}, \bibinfo{person}{Nachoem Wijnberg}, {and} \bibinfo{person}{Marcel Worring}.} \bibinfo{year}{2024}\natexlab{}.
\newblock \showarticletitle{Set2Seq Transformer: Temporal and Positional-Aware Set Representations for Sequential Multiple-Instance Learning}.
\newblock \bibinfo{journal}{\emph{arXiv preprint arXiv:2408.03404}} (\bibinfo{year}{2024}).
\newblock


\bibitem[Efthymiou et~al\mbox{.}(2021)]%
        {artsagenet}
\bibfield{author}{\bibinfo{person}{Athanasios Efthymiou}, \bibinfo{person}{Stevan Rudinac}, \bibinfo{person}{Monika Kackovic}, \bibinfo{person}{Marcel Worring}, {and} \bibinfo{person}{Nachoem Wijnberg}.} \bibinfo{year}{2021}\natexlab{}.
\newblock \showarticletitle{Graph Neural Networks for Knowledge Enhanced Visual Representation of Paintings}. In \bibinfo{booktitle}{\emph{Proceedings of the 29th ACM International Conference on Multimedia}} (Virtual Event, China) \emph{(\bibinfo{series}{MM '21})}. \bibinfo{publisher}{Association for Computing Machinery}, \bibinfo{address}{New York, NY, USA}, \bibinfo{pages}{3710–3719}.
\newblock
\showISBNx{9781450386517}
\href{https://doi.org/10.1145/3474085.3475586}{doi:\nolinkurl{10.1145/3474085.3475586}}


\bibitem[Egger et~al\mbox{.}(2024)]%
        {relik}
\bibfield{author}{\bibinfo{person}{Maximilian~K. Egger}, \bibinfo{person}{Wenyue Ma}, \bibinfo{person}{Davide Mottin}, \bibinfo{person}{Panagiotis Karras}, \bibinfo{person}{Ilaria Bordino}, \bibinfo{person}{Francesco Gullo}, {and} \bibinfo{person}{Aris Anagnostopoulos}.} \bibinfo{year}{2024}\natexlab{}.
\newblock \showarticletitle{ReliK: A Reliability Measure for Knowledge Graph Embeddings}. In \bibinfo{booktitle}{\emph{Proceedings of the ACM Web Conference 2024}} (Singapore, Singapore) \emph{(\bibinfo{series}{WWW '24})}. \bibinfo{publisher}{Association for Computing Machinery}, \bibinfo{address}{New York, NY, USA}, \bibinfo{pages}{2009–2019}.
\newblock
\showISBNx{9798400701719}
\href{https://doi.org/10.1145/3589334.3645430}{doi:\nolinkurl{10.1145/3589334.3645430}}


\bibitem[El~Vaigh et~al\mbox{.}(2021)]%
        {gcnboost}
\bibfield{author}{\bibinfo{person}{Cheikh~Brahim El~Vaigh}, \bibinfo{person}{Noa Garcia}, \bibinfo{person}{Benjamin Renoust}, \bibinfo{person}{Chenhui Chu}, \bibinfo{person}{Yuta Nakashima}, {and} \bibinfo{person}{Hajime Nagahara}.} \bibinfo{year}{2021}\natexlab{}.
\newblock \showarticletitle{GCNBoost: Artwork Classification by Label Propagation through a Knowledge Graph}. In \bibinfo{booktitle}{\emph{Proceedings of the 2021 International Conference on Multimedia Retrieval}} (Taipei, Taiwan) \emph{(\bibinfo{series}{ICMR '21})}. \bibinfo{publisher}{Association for Computing Machinery}, \bibinfo{address}{New York, NY, USA}, \bibinfo{pages}{92–100}.
\newblock
\showISBNx{9781450384636}
\href{https://doi.org/10.1145/3460426.3463636}{doi:\nolinkurl{10.1145/3460426.3463636}}


\bibitem[Elgammal et~al\mbox{.}(2018)]%
        {elgammal2018shape}
\bibfield{author}{\bibinfo{person}{Ahmed Elgammal}, \bibinfo{person}{Bingchen Liu}, \bibinfo{person}{Diana Kim}, \bibinfo{person}{Mohamed Elhoseiny}, {and} \bibinfo{person}{Marian Mazzone}.} \bibinfo{year}{2018}\natexlab{}.
\newblock \showarticletitle{The Shape of Art History in the Eyes of the Machine}.
\newblock \bibinfo{journal}{\emph{Proceedings of the AAAI Conference on Artificial Intelligence}} \bibinfo{volume}{32}, \bibinfo{number}{1} (\bibinfo{date}{Apr.} \bibinfo{year}{2018}).
\newblock
\href{https://doi.org/10.1609/aaai.v32i1.11894}{doi:\nolinkurl{10.1609/aaai.v32i1.11894}}


\bibitem[Galkin et~al\mbox{.}(2022)]%
        {nodepiece}
\bibfield{author}{\bibinfo{person}{Mikhail Galkin}, \bibinfo{person}{Etienne Denis}, \bibinfo{person}{Jiapeng Wu}, {and} \bibinfo{person}{William~L. Hamilton}.} \bibinfo{year}{2022}\natexlab{}.
\newblock \showarticletitle{NodePiece: Compositional and Parameter-Efficient Representations of Large Knowledge Graphs}. In \bibinfo{booktitle}{\emph{International Conference on Learning Representations}}.
\newblock
\urldef\tempurl%
\url{https://openreview.net/forum?id=xMJWUKJnFSw}
\showURL{%
\tempurl}


\bibitem[Gao et~al\mbox{.}(2025)]%
        {mckgc}
\bibfield{author}{\bibinfo{person}{Yuxiao Gao}, \bibinfo{person}{Fuwei Zhang}, \bibinfo{person}{Zhao Zhang}, \bibinfo{person}{Xiaoshuang Min}, {and} \bibinfo{person}{Fuzhen Zhuang}.} \bibinfo{year}{2025}\natexlab{}.
\newblock \showarticletitle{Mixed-Curvature Multi-Modal Knowledge Graph Completion}.
\newblock \bibinfo{journal}{\emph{Proceedings of the AAAI Conference on Artificial Intelligence}} \bibinfo{volume}{39}, \bibinfo{number}{11} (\bibinfo{date}{Apr.} \bibinfo{year}{2025}), \bibinfo{pages}{11699--11707}.
\newblock
\href{https://doi.org/10.1609/aaai.v39i11.33273}{doi:\nolinkurl{10.1609/aaai.v39i11.33273}}


\bibitem[Garcia et~al\mbox{.}(2020)]%
        {contextnet}
\bibfield{author}{\bibinfo{person}{Noa Garcia}, \bibinfo{person}{Benjamin Renoust}, {and} \bibinfo{person}{Yuta Nakashima}.} \bibinfo{year}{2020}\natexlab{}.
\newblock \showarticletitle{ContextNet: representation and exploration for painting classification and retrieval in context}.
\newblock \bibinfo{journal}{\emph{International Journal of Multimedia Information Retrieval}} \bibinfo{volume}{9}, \bibinfo{number}{1} (\bibinfo{year}{2020}), \bibinfo{pages}{17--30}.
\newblock


\bibitem[Gong et~al\mbox{.}(2025)]%
        {uknow}
\bibfield{author}{\bibinfo{person}{Biao Gong}, \bibinfo{person}{Shuai Tan}, \bibinfo{person}{Yutong Feng}, \bibinfo{person}{Xiaoying Xie}, \bibinfo{person}{Yuyuan Li}, \bibinfo{person}{Chaochao Chen}, \bibinfo{person}{Kecheng Zheng}, \bibinfo{person}{Yujun Shen}, {and} \bibinfo{person}{Deli Zhao}.} \bibinfo{year}{2025}\natexlab{}.
\newblock \showarticletitle{{UK}now: A Unified Knowledge Protocol with Multimodal Knowledge Graph Datasets for Reasoning and Vision-Language Pre-Training}. In \bibinfo{booktitle}{\emph{Proceedings of the 38th International Conference on Neural Information Processing Systems}} (Vancouver, BC, Canada) \emph{(\bibinfo{series}{NIPS '24})}. \bibinfo{publisher}{Curran Associates Inc.}, \bibinfo{address}{Red Hook, NY, USA}, Article \bibinfo{articleno}{306}, \bibinfo{numpages}{22}~pages.
\newblock
\showISBNx{9798331314385}


\bibitem[Goyal et~al\mbox{.}(2019)]%
        {vqa_2}
\bibfield{author}{\bibinfo{person}{Yash Goyal}, \bibinfo{person}{Tejas Khot}, \bibinfo{person}{Aishwarya Agrawal}, \bibinfo{person}{Douglas Summers-Stay}, \bibinfo{person}{Dhruv Batra}, {and} \bibinfo{person}{Devi Parikh}.} \bibinfo{year}{2019}\natexlab{}.
\newblock \showarticletitle{Making the V in VQA Matter: Elevating the Role of Image Understanding in Visual Question Answering}.
\newblock \bibinfo{journal}{\emph{Int. J. Comput. Vision}} \bibinfo{volume}{127}, \bibinfo{number}{4} (\bibinfo{date}{April} \bibinfo{year}{2019}), \bibinfo{pages}{398–414}.
\newblock
\showISSN{0920-5691}
\href{https://doi.org/10.1007/s11263-018-1116-0}{doi:\nolinkurl{10.1007/s11263-018-1116-0}}


\bibitem[Grover and Leskovec(2016)]%
        {node2vec}
\bibfield{author}{\bibinfo{person}{Aditya Grover} {and} \bibinfo{person}{Jure Leskovec}.} \bibinfo{year}{2016}\natexlab{}.
\newblock \showarticletitle{node2vec: Scalable Feature Learning for Networks}. In \bibinfo{booktitle}{\emph{Proceedings of the 22nd ACM SIGKDD International Conference on Knowledge Discovery and Data Mining}} (San Francisco, California, USA) \emph{(\bibinfo{series}{KDD '16})}. \bibinfo{publisher}{Association for Computing Machinery}, \bibinfo{address}{New York, NY, USA}, \bibinfo{pages}{855–864}.
\newblock
\showISBNx{9781450342322}
\href{https://doi.org/10.1145/2939672.2939754}{doi:\nolinkurl{10.1145/2939672.2939754}}


\bibitem[Hamilton et~al\mbox{.}(2017)]%
        {graphsage}
\bibfield{author}{\bibinfo{person}{Will Hamilton}, \bibinfo{person}{Zhitao Ying}, {and} \bibinfo{person}{Jure Leskovec}.} \bibinfo{year}{2017}\natexlab{}.
\newblock \showarticletitle{Inductive Representation Learning on Large Graphs}. In \bibinfo{booktitle}{\emph{Advances in Neural Information Processing Systems}}, \bibfield{editor}{\bibinfo{person}{I.~Guyon}, \bibinfo{person}{U.~Von Luxburg}, \bibinfo{person}{S.~Bengio}, \bibinfo{person}{H.~Wallach}, \bibinfo{person}{R.~Fergus}, \bibinfo{person}{S.~Vishwanathan}, {and} \bibinfo{person}{R.~Garnett}} (Eds.), Vol.~\bibinfo{volume}{30}. \bibinfo{publisher}{Curran Associates, Inc.}
\newblock
\urldef\tempurl%
\url{https://proceedings.neurips.cc/paper_files/paper/2017/file/5dd9db5e033da9c6fb5ba83c7a7ebea9-Paper.pdf}
\showURL{%
\tempurl}


\bibitem[He et~al\mbox{.}(2025)]%
        {unigraph_2}
\bibfield{author}{\bibinfo{person}{Yufei He}, \bibinfo{person}{Yuan Sui}, \bibinfo{person}{Xiaoxin He}, \bibinfo{person}{Yue Liu}, \bibinfo{person}{Yifei Sun}, {and} \bibinfo{person}{Bryan Hooi}.} \bibinfo{year}{2025}\natexlab{}.
\newblock \showarticletitle{UniGraph2: Learning a Unified Embedding Space to Bind Multimodal Graphs}. In \bibinfo{booktitle}{\emph{Proceedings of the ACM on Web Conference 2025}} (Sydney NSW, Australia) \emph{(\bibinfo{series}{WWW '25})}. \bibinfo{publisher}{Association for Computing Machinery}, \bibinfo{address}{New York, NY, USA}, \bibinfo{pages}{1759–1770}.
\newblock
\showISBNx{9798400712746}
\href{https://doi.org/10.1145/3696410.3714818}{doi:\nolinkurl{10.1145/3696410.3714818}}


\bibitem[Hu and Singh(2021)]%
        {unit}
\bibfield{author}{\bibinfo{person}{Ronghang Hu} {and} \bibinfo{person}{Amanpreet Singh}.} \bibinfo{year}{2021}\natexlab{}.
\newblock \showarticletitle{UniT: Multimodal Multitask Learning with a Unified Transformer}. In \bibinfo{booktitle}{\emph{2021 IEEE/CVF International Conference on Computer Vision (ICCV)}}. \bibinfo{pages}{1419--1429}.
\newblock
\href{https://doi.org/10.1109/ICCV48922.2021.00147}{doi:\nolinkurl{10.1109/ICCV48922.2021.00147}}


\bibitem[Hu et~al\mbox{.}(2021)]%
        {ogblsc}
\bibfield{author}{\bibinfo{person}{Weihua Hu}, \bibinfo{person}{Matthias Fey}, \bibinfo{person}{Hongyu Ren}, \bibinfo{person}{Maho Nakata}, \bibinfo{person}{Yuxiao Dong}, {and} \bibinfo{person}{Jure Leskovec}.} \bibinfo{year}{2021}\natexlab{}.
\newblock \showarticletitle{{OGB}-{LSC}: A Large-Scale Challenge for Machine Learning on Graphs}. In \bibinfo{booktitle}{\emph{Thirty-fifth Conference on Neural Information Processing Systems Datasets and Benchmarks Track (Round 2)}}.
\newblock
\urldef\tempurl%
\url{https://openreview.net/forum?id=qkcLxoC52kL}
\showURL{%
\tempurl}


\bibitem[Hu et~al\mbox{.}(2020)]%
        {ogb}
\bibfield{author}{\bibinfo{person}{Weihua Hu}, \bibinfo{person}{Matthias Fey}, \bibinfo{person}{Marinka Zitnik}, \bibinfo{person}{Yuxiao Dong}, \bibinfo{person}{Hongyu Ren}, \bibinfo{person}{Bowen Liu}, \bibinfo{person}{Michele Catasta}, {and} \bibinfo{person}{Jure Leskovec}.} \bibinfo{year}{2020}\natexlab{}.
\newblock \showarticletitle{Open Graph Benchmark: Datasets for Machine Learning on Graphs}. In \bibinfo{booktitle}{\emph{Advances in Neural Information Processing Systems}}, \bibfield{editor}{\bibinfo{person}{H.~Larochelle}, \bibinfo{person}{M.~Ranzato}, \bibinfo{person}{R.~Hadsell}, \bibinfo{person}{M.F. Balcan}, {and} \bibinfo{person}{H.~Lin}} (Eds.), Vol.~\bibinfo{volume}{33}. \bibinfo{publisher}{Curran Associates, Inc.}, \bibinfo{pages}{22118--22133}.
\newblock
\urldef\tempurl%
\url{https://proceedings.neurips.cc/paper_files/paper/2020/file/fb60d411a5c5b72b2e7d3527cfc84fd0-Paper.pdf}
\showURL{%
\tempurl}


\bibitem[Jia et~al\mbox{.}(2021)]%
        {align}
\bibfield{author}{\bibinfo{person}{Chao Jia}, \bibinfo{person}{Yinfei Yang}, \bibinfo{person}{Ye Xia}, \bibinfo{person}{Yi-Ting Chen}, \bibinfo{person}{Zarana Parekh}, \bibinfo{person}{Hieu Pham}, \bibinfo{person}{Quoc Le}, \bibinfo{person}{Yun-Hsuan Sung}, \bibinfo{person}{Zhen Li}, {and} \bibinfo{person}{Tom Duerig}.} \bibinfo{year}{2021}\natexlab{}.
\newblock \showarticletitle{Scaling Up Visual and Vision-Language Representation Learning With Noisy Text Supervision}. In \bibinfo{booktitle}{\emph{Proceedings of the 38th International Conference on Machine Learning}} \emph{(\bibinfo{series}{Proceedings of Machine Learning Research}, Vol.~\bibinfo{volume}{139})}, \bibfield{editor}{\bibinfo{person}{Marina Meila} {and} \bibinfo{person}{Tong Zhang}} (Eds.). \bibinfo{publisher}{PMLR}, \bibinfo{pages}{4904--4916}.
\newblock
\urldef\tempurl%
\url{https://proceedings.mlr.press/v139/jia21b.html}
\showURL{%
\tempurl}


\bibitem[Kim et~al\mbox{.}(2020)]%
        {mtl_kg}
\bibfield{author}{\bibinfo{person}{Bosung Kim}, \bibinfo{person}{Taesuk Hong}, \bibinfo{person}{Youngjoong Ko}, {and} \bibinfo{person}{Jungyun Seo}.} \bibinfo{year}{2020}\natexlab{}.
\newblock \showarticletitle{Multi-Task Learning for Knowledge Graph Completion with Pre-trained Language Models}. In \bibinfo{booktitle}{\emph{Proceedings of the 28th International Conference on Computational Linguistics}}, \bibfield{editor}{\bibinfo{person}{Donia Scott}, \bibinfo{person}{Nuria Bel}, {and} \bibinfo{person}{Chengqing Zong}} (Eds.). \bibinfo{publisher}{International Committee on Computational Linguistics}, \bibinfo{address}{Barcelona, Spain (Online)}, \bibinfo{pages}{1737--1743}.
\newblock
\href{https://doi.org/10.18653/v1/2020.coling-main.153}{doi:\nolinkurl{10.18653/v1/2020.coling-main.153}}


\bibitem[Lee et~al\mbox{.}(2023)]%
        {vista}
\bibfield{author}{\bibinfo{person}{Jaejun Lee}, \bibinfo{person}{Chanyoung Chung}, \bibinfo{person}{Hochang Lee}, \bibinfo{person}{Sungho Jo}, {and} \bibinfo{person}{Joyce~Jiyoung Whang}.} \bibinfo{year}{2023}\natexlab{}.
\newblock \showarticletitle{{VISTA}: Visual-Textual Knowledge Graph Representation Learning}. In \bibinfo{booktitle}{\emph{The 2023 Conference on Empirical Methods in Natural Language Processing}}.
\newblock
\urldef\tempurl%
\url{https://openreview.net/forum?id=Y2wUa9n7sr}
\showURL{%
\tempurl}


\bibitem[Lee et~al\mbox{.}(2024)]%
        {mr_mkg}
\bibfield{author}{\bibinfo{person}{Junlin Lee}, \bibinfo{person}{Yequan Wang}, \bibinfo{person}{Jing Li}, {and} \bibinfo{person}{Min Zhang}.} \bibinfo{year}{2024}\natexlab{}.
\newblock \showarticletitle{Multimodal Reasoning with Multimodal Knowledge Graph}. In \bibinfo{booktitle}{\emph{Proceedings of the 62nd Annual Meeting of the Association for Computational Linguistics (Volume 1: Long Papers)}}, \bibfield{editor}{\bibinfo{person}{Lun-Wei Ku}, \bibinfo{person}{Andre Martins}, {and} \bibinfo{person}{Vivek Srikumar}} (Eds.). \bibinfo{publisher}{Association for Computational Linguistics}, \bibinfo{address}{Bangkok, Thailand}, \bibinfo{pages}{10767--10782}.
\newblock
\href{https://doi.org/10.18653/v1/2024.acl-long.579}{doi:\nolinkurl{10.18653/v1/2024.acl-long.579}}


\bibitem[Li et~al\mbox{.}(2023)]%
        {blip_2}
\bibfield{author}{\bibinfo{person}{Junnan Li}, \bibinfo{person}{Dongxu Li}, \bibinfo{person}{Silvio Savarese}, {and} \bibinfo{person}{Steven Hoi}.} \bibinfo{year}{2023}\natexlab{}.
\newblock \showarticletitle{{BLIP}-2: Bootstrapping Language-Image Pre-training with Frozen Image Encoders and Large Language Models}. In \bibinfo{booktitle}{\emph{Proceedings of the 40th International Conference on Machine Learning}} \emph{(\bibinfo{series}{Proceedings of Machine Learning Research}, Vol.~\bibinfo{volume}{202})}, \bibfield{editor}{\bibinfo{person}{Andreas Krause}, \bibinfo{person}{Emma Brunskill}, \bibinfo{person}{Kyunghyun Cho}, \bibinfo{person}{Barbara Engelhardt}, \bibinfo{person}{Sivan Sabato}, {and} \bibinfo{person}{Jonathan Scarlett}} (Eds.). \bibinfo{publisher}{PMLR}, \bibinfo{pages}{19730--19742}.
\newblock
\urldef\tempurl%
\url{https://proceedings.mlr.press/v202/li23q.html}
\showURL{%
\tempurl}


\bibitem[Li et~al\mbox{.}(2022)]%
        {blip}
\bibfield{author}{\bibinfo{person}{Junnan Li}, \bibinfo{person}{Dongxu Li}, \bibinfo{person}{Caiming Xiong}, {and} \bibinfo{person}{Steven Hoi}.} \bibinfo{year}{2022}\natexlab{}.
\newblock \showarticletitle{{BLIP}: Bootstrapping Language-Image Pre-training for Unified Vision-Language Understanding and Generation}. In \bibinfo{booktitle}{\emph{Proceedings of the 39th International Conference on Machine Learning}} \emph{(\bibinfo{series}{Proceedings of Machine Learning Research}, Vol.~\bibinfo{volume}{162})}, \bibfield{editor}{\bibinfo{person}{Kamalika Chaudhuri}, \bibinfo{person}{Stefanie Jegelka}, \bibinfo{person}{Le~Song}, \bibinfo{person}{Csaba Szepesvari}, \bibinfo{person}{Gang Niu}, {and} \bibinfo{person}{Sivan Sabato}} (Eds.). \bibinfo{publisher}{PMLR}, \bibinfo{pages}{12888--12900}.
\newblock
\urldef\tempurl%
\url{https://proceedings.mlr.press/v162/li22n.html}
\showURL{%
\tempurl}


\bibitem[Liang et~al\mbox{.}(2024)]%
        {recpiece}
\bibfield{author}{\bibinfo{person}{Ke Liang}, \bibinfo{person}{Yue Liu}, \bibinfo{person}{Hao Li}, \bibinfo{person}{Lingyuan Meng}, \bibinfo{person}{Suyuan Liu}, \bibinfo{person}{Siwei Wang}, \bibinfo{person}{Sihang Zhou}, {and} \bibinfo{person}{Xinwang Liu}.} \bibinfo{year}{2024}\natexlab{}.
\newblock \showarticletitle{Clustering then Propagation: Select Better Anchors for Knowledge Graph Embedding}. In \bibinfo{booktitle}{\emph{Advances in Neural Information Processing Systems}}, \bibfield{editor}{\bibinfo{person}{A.~Globerson}, \bibinfo{person}{L.~Mackey}, \bibinfo{person}{D.~Belgrave}, \bibinfo{person}{A.~Fan}, \bibinfo{person}{U.~Paquet}, \bibinfo{person}{J.~Tomczak}, {and} \bibinfo{person}{C.~Zhang}} (Eds.), Vol.~\bibinfo{volume}{37}. \bibinfo{publisher}{Curran Associates, Inc.}, \bibinfo{pages}{9449--9473}.
\newblock
\urldef\tempurl%
\url{https://proceedings.neurips.cc/paper_files/paper/2024/file/12143893d9d37c3569dda800b95cabd9-Paper-Conference.pdf}
\showURL{%
\tempurl}


\bibitem[Liu et~al\mbox{.}(2019)]%
        {mmkg}
\bibfield{author}{\bibinfo{person}{Ye Liu}, \bibinfo{person}{Hui Li}, \bibinfo{person}{Alberto Garcia-Duran}, \bibinfo{person}{Mathias Niepert}, \bibinfo{person}{Daniel Onoro-Rubio}, {and} \bibinfo{person}{David~S. Rosenblum}.} \bibinfo{year}{2019}\natexlab{}.
\newblock \showarticletitle{MMKG: Multi-modal Knowledge Graphs}. In \bibinfo{booktitle}{\emph{The Semantic Web}}, \bibfield{editor}{\bibinfo{person}{Pascal Hitzler}, \bibinfo{person}{Miriam Fern{\'a}ndez}, \bibinfo{person}{Krzysztof Janowicz}, \bibinfo{person}{Amrapali Zaveri}, \bibinfo{person}{Alasdair~J.G. Gray}, \bibinfo{person}{Vanessa Lopez}, \bibinfo{person}{Armin Haller}, {and} \bibinfo{person}{Karl Hammar}} (Eds.). \bibinfo{publisher}{Springer International Publishing}, \bibinfo{address}{Cham}, \bibinfo{pages}{459--474}.
\newblock
\showISBNx{978-3-030-21348-0}


\bibitem[Lu et~al\mbox{.}(2019)]%
        {vilbert}
\bibfield{author}{\bibinfo{person}{Jiasen Lu}, \bibinfo{person}{Dhruv Batra}, \bibinfo{person}{Devi Parikh}, {and} \bibinfo{person}{Stefan Lee}.} \bibinfo{year}{2019}\natexlab{}.
\newblock \showarticletitle{ViLBERT: Pretraining Task-Agnostic Visiolinguistic Representations for Vision-and-Language Tasks}. In \bibinfo{booktitle}{\emph{Advances in Neural Information Processing Systems}}, \bibfield{editor}{\bibinfo{person}{H.~Wallach}, \bibinfo{person}{H.~Larochelle}, \bibinfo{person}{A.~Beygelzimer}, \bibinfo{person}{F.~d\textquotesingle Alch\'{e}-Buc}, \bibinfo{person}{E.~Fox}, {and} \bibinfo{person}{R.~Garnett}} (Eds.), Vol.~\bibinfo{volume}{32}. \bibinfo{publisher}{Curran Associates, Inc.}
\newblock
\urldef\tempurl%
\url{https://proceedings.neurips.cc/paper_files/paper/2019/file/c74d97b01eae257e44aa9d5bade97baf-Paper.pdf}
\showURL{%
\tempurl}


\bibitem[Lu et~al\mbox{.}(2022)]%
        {mmkrl}
\bibfield{author}{\bibinfo{person}{Xinyu Lu}, \bibinfo{person}{Lifang Wang}, \bibinfo{person}{Zejun Jiang}, \bibinfo{person}{Shichang He}, {and} \bibinfo{person}{Shizhong Liu}.} \bibinfo{year}{2022}\natexlab{}.
\newblock \showarticletitle{MMKRL: A robust embedding approach for multi-modal knowledge graph representation learning}.
\newblock \bibinfo{journal}{\emph{Applied Intelligence}} \bibinfo{volume}{52}, \bibinfo{number}{7} (\bibinfo{date}{May} \bibinfo{year}{2022}), \bibinfo{pages}{7480–7497}.
\newblock
\showISSN{0924-669X}
\href{https://doi.org/10.1007/s10489-021-02693-9}{doi:\nolinkurl{10.1007/s10489-021-02693-9}}


\bibitem[Lu et~al\mbox{.}(2024)]%
        {tcl}
\bibfield{author}{\bibinfo{person}{Yuxing Lu}, \bibinfo{person}{Weichen Zhao}, \bibinfo{person}{Nan Sun}, {and} \bibinfo{person}{Jinzhuo Wang}.} \bibinfo{year}{2024}\natexlab{}.
\newblock \showarticletitle{Enhancing Multimodal Knowledge Graph Representation Learning through Triple Contrastive Learning}. In \bibinfo{booktitle}{\emph{Proceedings of the Thirty-Third International Joint Conference on Artificial Intelligence, {IJCAI-24}}}, \bibfield{editor}{\bibinfo{person}{Kate Larson}} (Ed.). \bibinfo{publisher}{International Joint Conferences on Artificial Intelligence Organization}, \bibinfo{pages}{5963--5971}.
\newblock
\href{https://doi.org/10.24963/ijcai.2024/659}{doi:\nolinkurl{10.24963/ijcai.2024/659}}
\newblock
\shownote{Main Track}.


\bibitem[Mao et~al\mbox{.}(2017)]%
        {deepart}
\bibfield{author}{\bibinfo{person}{Hui Mao}, \bibinfo{person}{Ming Cheung}, {and} \bibinfo{person}{James She}.} \bibinfo{year}{2017}\natexlab{}.
\newblock \showarticletitle{DeepArt: Learning Joint Representations of Visual Arts}. In \bibinfo{booktitle}{\emph{Proceedings of the 25th ACM International Conference on Multimedia}} (Mountain View, California, USA) \emph{(\bibinfo{series}{MM '17})}. \bibinfo{publisher}{Association for Computing Machinery}, \bibinfo{address}{New York, NY, USA}, \bibinfo{pages}{1183–1191}.
\newblock
\showISBNx{9781450349062}
\href{https://doi.org/10.1145/3123266.3123405}{doi:\nolinkurl{10.1145/3123266.3123405}}


\bibitem[Miller(1995)]%
        {wordnet}
\bibfield{author}{\bibinfo{person}{George~A. Miller}.} \bibinfo{year}{1995}\natexlab{}.
\newblock \showarticletitle{WordNet: a lexical database for English}.
\newblock \bibinfo{journal}{\emph{Commun. ACM}} \bibinfo{volume}{38}, \bibinfo{number}{11} (\bibinfo{date}{Nov.} \bibinfo{year}{1995}), \bibinfo{pages}{39–41}.
\newblock
\showISSN{0001-0782}
\href{https://doi.org/10.1145/219717.219748}{doi:\nolinkurl{10.1145/219717.219748}}


\bibitem[Mohamed et~al\mbox{.}(2022)]%
        {artemis_v2}
\bibfield{author}{\bibinfo{person}{Youssef Mohamed}, \bibinfo{person}{Faizan~Farooq Khan}, \bibinfo{person}{Kilichbek Haydarov}, {and} \bibinfo{person}{Mohamed Elhoseiny}.} \bibinfo{year}{2022}\natexlab{}.
\newblock \showarticletitle{It is Okay to Not Be Okay: Overcoming Emotional Bias in Affective Image Captioning by Contrastive Data Collection}. In \bibinfo{booktitle}{\emph{2022 IEEE/CVF Conference on Computer Vision and Pattern Recognition (CVPR)}}. \bibinfo{pages}{21231--21240}.
\newblock
\href{https://doi.org/10.1109/CVPR52688.2022.02058}{doi:\nolinkurl{10.1109/CVPR52688.2022.02058}}


\bibitem[Morris et~al\mbox{.}(2020)]%
        {tu_dataset}
\bibfield{author}{\bibinfo{person}{Christopher Morris}, \bibinfo{person}{Nils~M. Kriege}, \bibinfo{person}{Franka Bause}, \bibinfo{person}{Kristian Kersting}, \bibinfo{person}{Petra Mutzel}, {and} \bibinfo{person}{Marion Neumann}.} \bibinfo{year}{2020}\natexlab{}.
\newblock \showarticletitle{TUDataset: A collection of benchmark datasets for learning with graphs}. In \bibinfo{booktitle}{\emph{ICML 2020 Workshop on Graph Representation Learning and Beyond (GRL+ 2020)}}.
\newblock
\showeprint[arxiv]{2007.08663}
\urldef\tempurl%
\url{www.graphlearning.io}
\showURL{%
\tempurl}


\bibitem[Mousselly-Sergieh et~al\mbox{.}(2018)]%
        {tbkge}
\bibfield{author}{\bibinfo{person}{Hatem Mousselly-Sergieh}, \bibinfo{person}{Teresa Botschen}, \bibinfo{person}{Iryna Gurevych}, {and} \bibinfo{person}{Stefan Roth}.} \bibinfo{year}{2018}\natexlab{}.
\newblock \showarticletitle{A Multimodal Translation-Based Approach for Knowledge Graph Representation Learning}. In \bibinfo{booktitle}{\emph{Proceedings of the Seventh Joint Conference on Lexical and Computational Semantics}}, \bibfield{editor}{\bibinfo{person}{Malvina Nissim}, \bibinfo{person}{Jonathan Berant}, {and} \bibinfo{person}{Alessandro Lenci}} (Eds.). \bibinfo{publisher}{Association for Computational Linguistics}, \bibinfo{address}{New Orleans, Louisiana}, \bibinfo{pages}{225--234}.
\newblock
\href{https://doi.org/10.18653/v1/S18-2027}{doi:\nolinkurl{10.18653/v1/S18-2027}}


\bibitem[Nguyen et~al\mbox{.}(2018)]%
        {conv_kb}
\bibfield{author}{\bibinfo{person}{Dai~Quoc Nguyen}, \bibinfo{person}{Tu~Dinh Nguyen}, \bibinfo{person}{Dat~Quoc Nguyen}, {and} \bibinfo{person}{Dinh Phung}.} \bibinfo{year}{2018}\natexlab{}.
\newblock \showarticletitle{A Novel Embedding Model for Knowledge Base Completion Based on Convolutional Neural Network}. In \bibinfo{booktitle}{\emph{Proceedings of the 2018 Conference of the North {A}merican Chapter of the Association for Computational Linguistics: Human Language Technologies, Volume 2 (Short Papers)}}, \bibfield{editor}{\bibinfo{person}{Marilyn Walker}, \bibinfo{person}{Heng Ji}, {and} \bibinfo{person}{Amanda Stent}} (Eds.). \bibinfo{publisher}{Association for Computational Linguistics}, \bibinfo{address}{New Orleans, Louisiana}, \bibinfo{pages}{327--333}.
\newblock
\href{https://doi.org/10.18653/v1/N18-2053}{doi:\nolinkurl{10.18653/v1/N18-2053}}


\bibitem[Nickel et~al\mbox{.}(2016)]%
        {hole}
\bibfield{author}{\bibinfo{person}{Maximilian Nickel}, \bibinfo{person}{Lorenzo Rosasco}, {and} \bibinfo{person}{Tomaso Poggio}.} \bibinfo{year}{2016}\natexlab{}.
\newblock \showarticletitle{Holographic Embeddings of Knowledge Graphs}.
\newblock \bibinfo{journal}{\emph{Proceedings of the AAAI Conference on Artificial Intelligence}} \bibinfo{volume}{30}, \bibinfo{number}{1} (\bibinfo{date}{Mar.} \bibinfo{year}{2016}).
\newblock
\href{https://doi.org/10.1609/aaai.v30i1.10314}{doi:\nolinkurl{10.1609/aaai.v30i1.10314}}


\bibitem[Nickel et~al\mbox{.}(2011)]%
        {rescal}
\bibfield{author}{\bibinfo{person}{Maximilian Nickel}, \bibinfo{person}{Volker Tresp}, {and} \bibinfo{person}{Hans-Peter Kriegel}.} \bibinfo{year}{2011}\natexlab{}.
\newblock \showarticletitle{A three-way model for collective learning on multi-relational data}. In \bibinfo{booktitle}{\emph{Proceedings of the 28th International Conference on International Conference on Machine Learning}} (Bellevue, Washington, USA) \emph{(\bibinfo{series}{ICML'11})}. \bibinfo{publisher}{Omnipress}, \bibinfo{address}{Madison, WI, USA}, \bibinfo{pages}{809–816}.
\newblock
\showISBNx{9781450306195}


\bibitem[Pan et~al\mbox{.}(2022)]%
        {knowledge_clip}
\bibfield{author}{\bibinfo{person}{Xuran Pan}, \bibinfo{person}{Tianzhu Ye}, \bibinfo{person}{Dongchen Han}, \bibinfo{person}{Shiji Song}, {and} \bibinfo{person}{Gao Huang}.} \bibinfo{year}{2022}\natexlab{}.
\newblock \showarticletitle{Contrastive Language-Image Pre-Training with Knowledge Graphs}. In \bibinfo{booktitle}{\emph{Advances in Neural Information Processing Systems}}, \bibfield{editor}{\bibinfo{person}{Alice~H. Oh}, \bibinfo{person}{Alekh Agarwal}, \bibinfo{person}{Danielle Belgrave}, {and} \bibinfo{person}{Kyunghyun Cho}} (Eds.).
\newblock
\urldef\tempurl%
\url{https://openreview.net/forum?id=4T3kbrzfeR}
\showURL{%
\tempurl}


\bibitem[Paszke et~al\mbox{.}(2019)]%
        {pytorch}
\bibfield{author}{\bibinfo{person}{Adam Paszke}, \bibinfo{person}{Sam Gross}, \bibinfo{person}{Francisco Massa}, \bibinfo{person}{Adam Lerer}, \bibinfo{person}{James Bradbury}, \bibinfo{person}{Gregory Chanan}, \bibinfo{person}{Trevor Killeen}, \bibinfo{person}{Zeming Lin}, \bibinfo{person}{Natalia Gimelshein}, \bibinfo{person}{Luca Antiga}, \bibinfo{person}{Alban Desmaison}, \bibinfo{person}{Andreas K\"{o}pf}, \bibinfo{person}{Edward Yang}, \bibinfo{person}{Zach DeVito}, \bibinfo{person}{Martin Raison}, \bibinfo{person}{Alykhan Tejani}, \bibinfo{person}{Sasank Chilamkurthy}, \bibinfo{person}{Benoit Steiner}, \bibinfo{person}{Lu Fang}, \bibinfo{person}{Junjie Bai}, {and} \bibinfo{person}{Soumith Chintala}.} \bibinfo{year}{2019}\natexlab{}.
\newblock \bibinfo{booktitle}{\emph{PyTorch: An Imperative Style, High-Performance Deep Learning Library}}.
\newblock \bibinfo{publisher}{Curran Associates Inc.}, \bibinfo{address}{Red Hook, NY, USA}.
\newblock


\bibitem[Perozzi et~al\mbox{.}(2014)]%
        {deepwalk}
\bibfield{author}{\bibinfo{person}{Bryan Perozzi}, \bibinfo{person}{Rami Al-Rfou}, {and} \bibinfo{person}{Steven Skiena}.} \bibinfo{year}{2014}\natexlab{}.
\newblock \showarticletitle{DeepWalk: online learning of social representations}. In \bibinfo{booktitle}{\emph{Proceedings of the 20th ACM SIGKDD International Conference on Knowledge Discovery and Data Mining}} (New York, New York, USA) \emph{(\bibinfo{series}{KDD '14})}. \bibinfo{publisher}{Association for Computing Machinery}, \bibinfo{address}{New York, NY, USA}, \bibinfo{pages}{701–710}.
\newblock
\showISBNx{9781450329569}
\href{https://doi.org/10.1145/2623330.2623732}{doi:\nolinkurl{10.1145/2623330.2623732}}


\bibitem[Radford et~al\mbox{.}(2021)]%
        {clip}
\bibfield{author}{\bibinfo{person}{Alec Radford}, \bibinfo{person}{Jong~Wook Kim}, \bibinfo{person}{Chris Hallacy}, \bibinfo{person}{Aditya Ramesh}, \bibinfo{person}{Gabriel Goh}, \bibinfo{person}{Sandhini Agarwal}, \bibinfo{person}{Girish Sastry}, \bibinfo{person}{Amanda Askell}, \bibinfo{person}{Pamela Mishkin}, \bibinfo{person}{Jack Clark}, {et~al\mbox{.}}} \bibinfo{year}{2021}\natexlab{}.
\newblock \showarticletitle{Learning transferable visual models from natural language supervision}. In \bibinfo{booktitle}{\emph{International conference on machine learning}}. PMLR, \bibinfo{pages}{8748--8763}.
\newblock


\bibitem[Ruffinelli et~al\mbox{.}(2020)]%
        {training_kges}
\bibfield{author}{\bibinfo{person}{Daniel Ruffinelli}, \bibinfo{person}{Samuel Broscheit}, {and} \bibinfo{person}{Rainer Gemulla}.} \bibinfo{year}{2020}\natexlab{}.
\newblock \showarticletitle{You CAN Teach an Old Dog New Tricks! On Training Knowledge Graph Embeddings}. In \bibinfo{booktitle}{\emph{International Conference on Learning Representations}}.
\newblock
\urldef\tempurl%
\url{https://openreview.net/forum?id=BkxSmlBFvr}
\showURL{%
\tempurl}


\bibitem[Scaringi et~al\mbox{.}(2025)]%
        {graphclip}
\bibfield{author}{\bibinfo{person}{Raffaele Scaringi}, \bibinfo{person}{Giuseppe Fiameni}, \bibinfo{person}{Gennaro Vessio}, {and} \bibinfo{person}{Giovanna Castellano}.} \bibinfo{year}{2025}\natexlab{}.
\newblock \showarticletitle{GraphCLIP: Image-graph contrastive learning for multimodal artwork classification}.
\newblock \bibinfo{journal}{\emph{Knowledge-Based Systems}}  \bibinfo{volume}{310} (\bibinfo{year}{2025}), \bibinfo{pages}{112857}.
\newblock
\showISSN{0950-7051}
\href{https://doi.org/10.1016/j.knosys.2024.112857}{doi:\nolinkurl{10.1016/j.knosys.2024.112857}}


\bibitem[Sen et~al\mbox{.}(2008)]%
        {cora}
\bibfield{author}{\bibinfo{person}{Prithviraj Sen}, \bibinfo{person}{Galileo Namata}, \bibinfo{person}{Mustafa Bilgic}, \bibinfo{person}{Lise Getoor}, \bibinfo{person}{Brian Galligher}, {and} \bibinfo{person}{Tina Eliassi-Rad}.} \bibinfo{year}{2008}\natexlab{}.
\newblock \showarticletitle{Collective Classification in Network Data}.
\newblock \bibinfo{journal}{\emph{AI Magazine}} \bibinfo{volume}{29}, \bibinfo{number}{3} (\bibinfo{date}{Sep.} \bibinfo{year}{2008}), \bibinfo{pages}{93}.
\newblock
\href{https://doi.org/10.1609/aimag.v29i3.2157}{doi:\nolinkurl{10.1609/aimag.v29i3.2157}}


\bibitem[Shang et~al\mbox{.}(2024)]%
        {lafa}
\bibfield{author}{\bibinfo{person}{Bin Shang}, \bibinfo{person}{Yinliang Zhao}, \bibinfo{person}{Jun Liu}, {and} \bibinfo{person}{Di Wang}.} \bibinfo{year}{2024}\natexlab{}.
\newblock \showarticletitle{LAFA: Multimodal Knowledge Graph Completion with Link Aware Fusion and Aggregation}.
\newblock \bibinfo{journal}{\emph{Proceedings of the AAAI Conference on Artificial Intelligence}} \bibinfo{volume}{38}, \bibinfo{number}{8} (\bibinfo{date}{Mar.} \bibinfo{year}{2024}), \bibinfo{pages}{8957--8965}.
\newblock
\href{https://doi.org/10.1609/aaai.v38i8.28744}{doi:\nolinkurl{10.1609/aaai.v38i8.28744}}


\bibitem[Singh et~al\mbox{.}(2022)]%
        {flava}
\bibfield{author}{\bibinfo{person}{Amanpreet Singh}, \bibinfo{person}{Ronghang Hu}, \bibinfo{person}{Vedanuj Goswami}, \bibinfo{person}{Guillaume Couairon}, \bibinfo{person}{Wojciech Galuba}, \bibinfo{person}{Marcus Rohrbach}, {and} \bibinfo{person}{Douwe Kiela}.} \bibinfo{year}{2022}\natexlab{}.
\newblock \showarticletitle{FLAVA: A Foundational Language and Vision Alignment Model}. In \bibinfo{booktitle}{\emph{Proceedings of the IEEE/CVF Conference on Computer Vision and Pattern Recognition (CVPR)}}. \bibinfo{pages}{15638--15650}.
\newblock


\bibitem[Strezoski and Worring(2018)]%
        {strezoski_omniart}
\bibfield{author}{\bibinfo{person}{Gjorgji Strezoski} {and} \bibinfo{person}{Marcel Worring}.} \bibinfo{year}{2018}\natexlab{}.
\newblock \showarticletitle{OmniArt: A Large-Scale Artistic Benchmark}.
\newblock \bibinfo{journal}{\emph{ACM Trans. Multimedia Comput. Commun. Appl.}} \bibinfo{volume}{14}, \bibinfo{number}{4}, Article \bibinfo{articleno}{88} (\bibinfo{date}{Oct.} \bibinfo{year}{2018}), \bibinfo{numpages}{21}~pages.
\newblock
\showISSN{1551-6857}
\href{https://doi.org/10.1145/3273022}{doi:\nolinkurl{10.1145/3273022}}


\bibitem[Sun et~al\mbox{.}(2019)]%
        {rotate}
\bibfield{author}{\bibinfo{person}{Zhiqing Sun}, \bibinfo{person}{Zhi-Hong Deng}, \bibinfo{person}{Jian-Yun Nie}, {and} \bibinfo{person}{Jian Tang}.} \bibinfo{year}{2019}\natexlab{}.
\newblock \showarticletitle{RotatE: Knowledge Graph Embedding by Relational Rotation in Complex Space}. In \bibinfo{booktitle}{\emph{International Conference on Learning Representations}}.
\newblock
\urldef\tempurl%
\url{https://openreview.net/forum?id=HkgEQnRqYQ}
\showURL{%
\tempurl}


\bibitem[Tan et~al\mbox{.}(2016)]%
        {tan_artgan}
\bibfield{author}{\bibinfo{person}{Wei~Ren Tan}, \bibinfo{person}{Chee~Seng Chan}, \bibinfo{person}{Hern{\'a}n~E Aguirre}, {and} \bibinfo{person}{Kiyoshi Tanaka}.} \bibinfo{year}{2016}\natexlab{}.
\newblock \showarticletitle{Ceci n'est pas une pipe: A deep convolutional network for fine-art paintings classification}. In \bibinfo{booktitle}{\emph{2016 IEEE international conference on image processing (ICIP)}}. IEEE, \bibinfo{pages}{3703--3707}.
\newblock


\bibitem[Tang et~al\mbox{.}(2019)]%
        {mkrl}
\bibfield{author}{\bibinfo{person}{Xing Tang}, \bibinfo{person}{Ling Chen}, \bibinfo{person}{Jun Cui}, {and} \bibinfo{person}{Baogang Wei}.} \bibinfo{year}{2019}\natexlab{}.
\newblock \showarticletitle{Knowledge representation learning with entity descriptions, hierarchical types, and textual relations}.
\newblock \bibinfo{journal}{\emph{Information Processing \& Management}} \bibinfo{volume}{56}, \bibinfo{number}{3} (\bibinfo{year}{2019}), \bibinfo{pages}{809--822}.
\newblock
\showISSN{0306-4573}
\href{https://doi.org/10.1016/j.ipm.2019.01.005}{doi:\nolinkurl{10.1016/j.ipm.2019.01.005}}


\bibitem[Trouillon et~al\mbox{.}(2016)]%
        {complex}
\bibfield{author}{\bibinfo{person}{Théo Trouillon}, \bibinfo{person}{Johannes Welbl}, \bibinfo{person}{Sebastian Riedel}, \bibinfo{person}{Eric Gaussier}, {and} \bibinfo{person}{Guillaume Bouchard}.} \bibinfo{year}{2016}\natexlab{}.
\newblock \showarticletitle{Complex Embeddings for Simple Link Prediction}. In \bibinfo{booktitle}{\emph{Proceedings of The 33rd International Conference on Machine Learning}} \emph{(\bibinfo{series}{Proceedings of Machine Learning Research}, Vol.~\bibinfo{volume}{48})}, \bibfield{editor}{\bibinfo{person}{Maria~Florina Balcan} {and} \bibinfo{person}{Kilian~Q. Weinberger}} (Eds.). \bibinfo{publisher}{PMLR}, \bibinfo{address}{New York, New York, USA}, \bibinfo{pages}{2071--2080}.
\newblock
\urldef\tempurl%
\url{https://proceedings.mlr.press/v48/trouillon16.html}
\showURL{%
\tempurl}


\bibitem[Tsimpoukelli et~al\mbox{.}(2021)]%
        {frozen}
\bibfield{author}{\bibinfo{person}{Maria Tsimpoukelli}, \bibinfo{person}{Jacob~L Menick}, \bibinfo{person}{Serkan Cabi}, \bibinfo{person}{S.~M.~Ali Eslami}, \bibinfo{person}{Oriol Vinyals}, {and} \bibinfo{person}{Felix Hill}.} \bibinfo{year}{2021}\natexlab{}.
\newblock \showarticletitle{Multimodal Few-Shot Learning with Frozen Language Models}. In \bibinfo{booktitle}{\emph{Advances in Neural Information Processing Systems}}, \bibfield{editor}{\bibinfo{person}{M.~Ranzato}, \bibinfo{person}{A.~Beygelzimer}, \bibinfo{person}{Y.~Dauphin}, \bibinfo{person}{P.S. Liang}, {and} \bibinfo{person}{J.~Wortman Vaughan}} (Eds.), Vol.~\bibinfo{volume}{34}. \bibinfo{publisher}{Curran Associates, Inc.}, \bibinfo{pages}{200--212}.
\newblock
\urldef\tempurl%
\url{https://proceedings.neurips.cc/paper_files/paper/2021/file/01b7575c38dac42f3cfb7d500438b875-Paper.pdf}
\showURL{%
\tempurl}


\bibitem[Usmani et~al\mbox{.}(2023)]%
        {towards_mkgs_data_spaces}
\bibfield{author}{\bibinfo{person}{Atiya Usmani}, \bibinfo{person}{M.~Jaleed Khan}, \bibinfo{person}{John G.~Breslin}, {and} \bibinfo{person}{Edward Curry}.} \bibinfo{year}{2023}\natexlab{}.
\newblock \showarticletitle{Towards Multimodal Knowledge Graphs for Data Spaces}. In \bibinfo{booktitle}{\emph{Companion Proceedings of the ACM Web Conference 2023}} (Austin, TX, USA) \emph{(\bibinfo{series}{WWW '23 Companion})}. \bibinfo{publisher}{Association for Computing Machinery}, \bibinfo{address}{New York, NY, USA}, \bibinfo{pages}{1494–1499}.
\newblock
\showISBNx{9781450394192}
\href{https://doi.org/10.1145/3543873.3587665}{doi:\nolinkurl{10.1145/3543873.3587665}}


\bibitem[Wang et~al\mbox{.}(2022a)]%
        {mkgrl_ms}
\bibfield{author}{\bibinfo{person}{Enqiang Wang}, \bibinfo{person}{Qing Yu}, \bibinfo{person}{Yelin Chen}, \bibinfo{person}{Wushouer Slamu}, {and} \bibinfo{person}{Xukang Luo}.} \bibinfo{year}{2022}\natexlab{a}.
\newblock \showarticletitle{Multi-modal knowledge graphs representation learning via multi-headed self-attention}.
\newblock \bibinfo{journal}{\emph{Information Fusion}}  \bibinfo{volume}{88} (\bibinfo{year}{2022}), \bibinfo{pages}{78--85}.
\newblock
\showISSN{1566-2535}
\href{https://doi.org/10.1016/j.inffus.2022.07.008}{doi:\nolinkurl{10.1016/j.inffus.2022.07.008}}


\bibitem[Wang et~al\mbox{.}(2024)]%
        {ime}
\bibfield{author}{\bibinfo{person}{Jiapu Wang}, \bibinfo{person}{Zheng Cui}, \bibinfo{person}{Boyue Wang}, \bibinfo{person}{Shirui Pan}, \bibinfo{person}{Junbin Gao}, \bibinfo{person}{Baocai Yin}, {and} \bibinfo{person}{Wen Gao}.} \bibinfo{year}{2024}\natexlab{}.
\newblock \showarticletitle{IME: Integrating Multi-curvature Shared and Specific Embedding for Temporal Knowledge Graph Completion}. In \bibinfo{booktitle}{\emph{Proceedings of the ACM Web Conference 2024}} (Singapore, Singapore) \emph{(\bibinfo{series}{WWW '24})}. \bibinfo{publisher}{Association for Computing Machinery}, \bibinfo{address}{New York, NY, USA}, \bibinfo{pages}{1954–1962}.
\newblock
\showISBNx{9798400701719}
\href{https://doi.org/10.1145/3589334.3645361}{doi:\nolinkurl{10.1145/3589334.3645361}}


\bibitem[Wang et~al\mbox{.}(2020)]%
        {mag_dataset}
\bibfield{author}{\bibinfo{person}{Kuansan Wang}, \bibinfo{person}{Zhihong Shen}, \bibinfo{person}{Chiyuan Huang}, \bibinfo{person}{Chieh-Han Wu}, \bibinfo{person}{Yuxiao Dong}, {and} \bibinfo{person}{Anshul Kanakia}.} \bibinfo{year}{2020}\natexlab{}.
\newblock \showarticletitle{Microsoft Academic Graph: When experts are not enough}.
\newblock \bibinfo{journal}{\emph{Quantitative Science Studies}} \bibinfo{volume}{1}, \bibinfo{number}{1} (\bibinfo{date}{02} \bibinfo{year}{2020}), \bibinfo{pages}{396--413}.
\newblock
\showISSN{2641-3337}
\showeprint{https://direct.mit.edu/qss/article-pdf/1/1/396/1760880/qss\_a\_00021.pdf}
\href{https://doi.org/10.1162/qss_a_00021}{doi:\nolinkurl{10.1162/qss_a_00021}}


\bibitem[Wang et~al\mbox{.}(2021)]%
        {rsme}
\bibfield{author}{\bibinfo{person}{Meng Wang}, \bibinfo{person}{Sen Wang}, \bibinfo{person}{Han Yang}, \bibinfo{person}{Zheng Zhang}, \bibinfo{person}{Xi Chen}, {and} \bibinfo{person}{Guilin Qi}.} \bibinfo{year}{2021}\natexlab{}.
\newblock \showarticletitle{Is Visual Context Really Helpful for Knowledge Graph? A Representation Learning Perspective}. In \bibinfo{booktitle}{\emph{Proceedings of the 29th ACM International Conference on Multimedia}} (Virtual Event, China) \emph{(\bibinfo{series}{MM '21})}. \bibinfo{publisher}{Association for Computing Machinery}, \bibinfo{address}{New York, NY, USA}, \bibinfo{pages}{2735–2743}.
\newblock
\showISBNx{9781450386517}
\href{https://doi.org/10.1145/3474085.3475470}{doi:\nolinkurl{10.1145/3474085.3475470}}


\bibitem[Wang et~al\mbox{.}(2025)]%
        {artrag}
\bibfield{author}{\bibinfo{person}{Shuai Wang}, \bibinfo{person}{Ivona Najdenkoska}, \bibinfo{person}{Hongyi Zhu}, \bibinfo{person}{Stevan Rudinac}, \bibinfo{person}{Monika Kackovic}, \bibinfo{person}{Nachoem Wijnberg}, {and} \bibinfo{person}{Marcel Worring}.} \bibinfo{year}{2025}\natexlab{}.
\newblock \showarticletitle{ArtRAG: Retrieval-Augmented Generation with Structured Context for Visual Art Understanding}. In \bibinfo{booktitle}{\emph{Proceedings of the 33rd ACM International Conference on Multimedia}} (Dublin, Ireland) \emph{(\bibinfo{series}{MM '25})}. \bibinfo{publisher}{Association for Computing Machinery}, \bibinfo{address}{New York, NY, USA}, \bibinfo{pages}{6700–6709}.
\newblock
\showISBNx{9798400720352}
\href{https://doi.org/10.1145/3746027.3755673}{doi:\nolinkurl{10.1145/3746027.3755673}}


\bibitem[Wang and Shen(2025)]%
        {otmkgrl}
\bibfield{author}{\bibinfo{person}{Tao Wang} {and} \bibinfo{person}{Bo Shen}.} \bibinfo{year}{2025}\natexlab{}.
\newblock \showarticletitle{OTMKGRL: a universal multimodal knowledge graph representation learning framework using optimal transport and cross-modal relation}.
\newblock \bibinfo{journal}{\emph{Applied Intelligence}} \bibinfo{volume}{55}, \bibinfo{number}{7} (\bibinfo{date}{March} \bibinfo{year}{2025}), \bibinfo{numpages}{20}~pages.
\newblock
\showISSN{0924-669X}
\href{https://doi.org/10.1007/s10489-025-06459-5}{doi:\nolinkurl{10.1007/s10489-025-06459-5}}


\bibitem[Wang et~al\mbox{.}(2023)]%
        {vqa_gnn}
\bibfield{author}{\bibinfo{person}{Yanan Wang}, \bibinfo{person}{Michihiro Yasunaga}, \bibinfo{person}{Hongyu Ren}, \bibinfo{person}{Shinya Wada}, {and} \bibinfo{person}{Jure Leskovec}.} \bibinfo{year}{2023}\natexlab{}.
\newblock \showarticletitle{VQA-GNN: Reasoning with Multimodal Knowledge via Graph Neural Networks for Visual Question Answering}. In \bibinfo{booktitle}{\emph{2023 IEEE/CVF International Conference on Computer Vision (ICCV)}}. \bibinfo{pages}{21525--21535}.
\newblock
\href{https://doi.org/10.1109/ICCV51070.2023.01973}{doi:\nolinkurl{10.1109/ICCV51070.2023.01973}}


\bibitem[Wang et~al\mbox{.}(2019)]%
        {transae}
\bibfield{author}{\bibinfo{person}{Zikang Wang}, \bibinfo{person}{Linjing Li}, \bibinfo{person}{Qiudan Li}, {and} \bibinfo{person}{Daniel Zeng}.} \bibinfo{year}{2019}\natexlab{}.
\newblock \showarticletitle{Multimodal Data Enhanced Representation Learning for Knowledge Graphs}. In \bibinfo{booktitle}{\emph{2019 International Joint Conference on Neural Networks (IJCNN)}}. \bibinfo{pages}{1--8}.
\newblock
\href{https://doi.org/10.1109/IJCNN.2019.8852079}{doi:\nolinkurl{10.1109/IJCNN.2019.8852079}}


\bibitem[Wang et~al\mbox{.}(2022b)]%
        {simvlm}
\bibfield{author}{\bibinfo{person}{Zirui Wang}, \bibinfo{person}{Jiahui Yu}, \bibinfo{person}{Adams~Wei Yu}, \bibinfo{person}{Zihang Dai}, \bibinfo{person}{Yulia Tsvetkov}, {and} \bibinfo{person}{Yuan Cao}.} \bibinfo{year}{2022}\natexlab{b}.
\newblock \showarticletitle{Sim{VLM}: Simple Visual Language Model Pretraining with Weak Supervision}. In \bibinfo{booktitle}{\emph{International Conference on Learning Representations}}.
\newblock
\urldef\tempurl%
\url{https://openreview.net/forum?id=GUrhfTuf_3}
\showURL{%
\tempurl}


\bibitem[WikiArt({[n.\,d.]})]%
        {wikiart}
\bibfield{author}{\bibinfo{person}{WikiArt}.} \bibinfo{year}{[n.\,d.]}\natexlab{}.
\newblock \bibinfo{title}{Visual Art Encyclopedia}.
\newblock
\urldef\tempurl%
\url{https://www.wikiart.org/}
\showURL{%
\tempurl}


\bibitem[Wilber et~al\mbox{.}(2017)]%
        {wilber_bam}
\bibfield{author}{\bibinfo{person}{Michael~J. Wilber}, \bibinfo{person}{Chen Fang}, \bibinfo{person}{Hailin Jin}, \bibinfo{person}{Aaron Hertzmann}, \bibinfo{person}{John Collomosse}, {and} \bibinfo{person}{Serge Belongie}.} \bibinfo{year}{2017}\natexlab{}.
\newblock \showarticletitle{BAM! The Behance Artistic Media Dataset for Recognition Beyond Photography}. In \bibinfo{booktitle}{\emph{The IEEE International Conference on Computer Vision (ICCV)}}.
\newblock


\bibitem[Xie et~al\mbox{.}(2020)]%
        {ikrl}
\bibfield{author}{\bibinfo{person}{Ruobing Xie}, \bibinfo{person}{Stefan Heinrich}, \bibinfo{person}{Zhiyuan Liu}, \bibinfo{person}{Cornelius Weber}, \bibinfo{person}{Yuan Yao}, \bibinfo{person}{Stefan Wermter}, {and} \bibinfo{person}{Maosong Sun}.} \bibinfo{year}{2020}\natexlab{}.
\newblock \showarticletitle{Integrating Image-Based and Knowledge-Based Representation Learning}.
\newblock \bibinfo{journal}{\emph{IEEE Transactions on Cognitive and Developmental Systems}} \bibinfo{volume}{12}, \bibinfo{number}{2} (\bibinfo{year}{2020}), \bibinfo{pages}{169--178}.
\newblock
\href{https://doi.org/10.1109/TCDS.2019.2906685}{doi:\nolinkurl{10.1109/TCDS.2019.2906685}}


\bibitem[Xie et~al\mbox{.}(2016)]%
        {dkrl}
\bibfield{author}{\bibinfo{person}{Ruobing Xie}, \bibinfo{person}{Zhiyuan Liu}, \bibinfo{person}{Jia Jia}, \bibinfo{person}{Huanbo Luan}, {and} \bibinfo{person}{Maosong Sun}.} \bibinfo{year}{2016}\natexlab{}.
\newblock \showarticletitle{Representation Learning of Knowledge Graphs with Entity Descriptions}.
\newblock \bibinfo{journal}{\emph{Proceedings of the AAAI Conference on Artificial Intelligence}} \bibinfo{volume}{30}, \bibinfo{number}{1} (\bibinfo{date}{Mar.} \bibinfo{year}{2016}).
\newblock
\href{https://doi.org/10.1609/aaai.v30i1.10329}{doi:\nolinkurl{10.1609/aaai.v30i1.10329}}


\bibitem[Xie et~al\mbox{.}(2017)]%
        {wn9_img_dataset}
\bibfield{author}{\bibinfo{person}{Ruobing Xie}, \bibinfo{person}{Zhiyuan Liu}, \bibinfo{person}{Huanbo Luan}, {and} \bibinfo{person}{Maosong Sun}.} \bibinfo{year}{2017}\natexlab{}.
\newblock \showarticletitle{Image-embodied Knowledge Representation Learning}. In \bibinfo{booktitle}{\emph{Proceedings of the Twenty-Sixth International Joint Conference on Artificial Intelligence, {IJCAI-17}}}. \bibinfo{pages}{3140--3146}.
\newblock
\href{https://doi.org/10.24963/ijcai.2017/438}{doi:\nolinkurl{10.24963/ijcai.2017/438}}


\bibitem[Xu et~al\mbox{.}(2022)]%
        {mmrns}
\bibfield{author}{\bibinfo{person}{Derong Xu}, \bibinfo{person}{Tong Xu}, \bibinfo{person}{Shiwei Wu}, \bibinfo{person}{Jingbo Zhou}, {and} \bibinfo{person}{Enhong Chen}.} \bibinfo{year}{2022}\natexlab{}.
\newblock \showarticletitle{Relation-enhanced Negative Sampling for Multimodal Knowledge Graph Completion}. In \bibinfo{booktitle}{\emph{Proceedings of the 30th ACM International Conference on Multimedia}} (Lisboa, Portugal) \emph{(\bibinfo{series}{MM '22})}. \bibinfo{publisher}{Association for Computing Machinery}, \bibinfo{address}{New York, NY, USA}, \bibinfo{pages}{3857–3866}.
\newblock
\showISBNx{9781450392037}
\href{https://doi.org/10.1145/3503161.3548388}{doi:\nolinkurl{10.1145/3503161.3548388}}


\bibitem[Yan et~al\mbox{.}(2024)]%
        {urban_clip}
\bibfield{author}{\bibinfo{person}{Yibo Yan}, \bibinfo{person}{Haomin Wen}, \bibinfo{person}{Siru Zhong}, \bibinfo{person}{Wei Chen}, \bibinfo{person}{Haodong Chen}, \bibinfo{person}{Qingsong Wen}, \bibinfo{person}{Roger Zimmermann}, {and} \bibinfo{person}{Yuxuan Liang}.} \bibinfo{year}{2024}\natexlab{}.
\newblock \showarticletitle{UrbanCLIP: Learning Text-enhanced Urban Region Profiling with Contrastive Language-Image Pretraining from the Web}. In \bibinfo{booktitle}{\emph{Proceedings of the ACM Web Conference 2024}} (Singapore, Singapore) \emph{(\bibinfo{series}{WWW '24})}. \bibinfo{publisher}{Association for Computing Machinery}, \bibinfo{address}{New York, NY, USA}, \bibinfo{pages}{4006–4017}.
\newblock
\showISBNx{9798400701719}
\href{https://doi.org/10.1145/3589334.3645378}{doi:\nolinkurl{10.1145/3589334.3645378}}


\bibitem[Yanardag and Vishwanathan(2015)]%
        {deep_graph_kernels}
\bibfield{author}{\bibinfo{person}{Pinar Yanardag} {and} \bibinfo{person}{S.V.N. Vishwanathan}.} \bibinfo{year}{2015}\natexlab{}.
\newblock \showarticletitle{Deep Graph Kernels}. In \bibinfo{booktitle}{\emph{Proceedings of the 21th ACM SIGKDD International Conference on Knowledge Discovery and Data Mining}} (Sydney, NSW, Australia) \emph{(\bibinfo{series}{KDD '15})}. \bibinfo{publisher}{Association for Computing Machinery}, \bibinfo{address}{New York, NY, USA}, \bibinfo{pages}{1365–1374}.
\newblock
\showISBNx{9781450336642}
\href{https://doi.org/10.1145/2783258.2783417}{doi:\nolinkurl{10.1145/2783258.2783417}}


\bibitem[Yang et~al\mbox{.}(2015)]%
        {distmult}
\bibfield{author}{\bibinfo{person}{Bishan Yang}, \bibinfo{person}{Wen{-}tau Yih}, \bibinfo{person}{Xiaodong He}, \bibinfo{person}{Jianfeng Gao}, {and} \bibinfo{person}{Li Deng}.} \bibinfo{year}{2015}\natexlab{}.
\newblock \showarticletitle{Embedding Entities and Relations for Learning and Inference in Knowledge Bases}. In \bibinfo{booktitle}{\emph{3rd International Conference on Learning Representations, {ICLR} 2015, San Diego, CA, USA, May 7-9, 2015, Conference Track Proceedings}}, \bibfield{editor}{\bibinfo{person}{Yoshua Bengio} {and} \bibinfo{person}{Yann LeCun}} (Eds.).
\newblock
\urldef\tempurl%
\url{http://arxiv.org/abs/1412.6575}
\showURL{%
\tempurl}


\bibitem[Yang et~al\mbox{.}(2016)]%
        {cora_v2}
\bibfield{author}{\bibinfo{person}{Zhilin Yang}, \bibinfo{person}{William Cohen}, {and} \bibinfo{person}{Ruslan Salakhudinov}.} \bibinfo{year}{2016}\natexlab{}.
\newblock \showarticletitle{Revisiting Semi-Supervised Learning with Graph Embeddings}. In \bibinfo{booktitle}{\emph{Proceedings of The 33rd International Conference on Machine Learning}} \emph{(\bibinfo{series}{Proceedings of Machine Learning Research}, Vol.~\bibinfo{volume}{48})}, \bibfield{editor}{\bibinfo{person}{Maria~Florina Balcan} {and} \bibinfo{person}{Kilian~Q. Weinberger}} (Eds.). \bibinfo{publisher}{PMLR}, \bibinfo{address}{New York, New York, USA}, \bibinfo{pages}{40--48}.
\newblock
\urldef\tempurl%
\url{https://proceedings.mlr.press/v48/yanga16.html}
\showURL{%
\tempurl}


\bibitem[Yao et~al\mbox{.}(2019)]%
        {kg_bert}
\bibfield{author}{\bibinfo{person}{Liang Yao}, \bibinfo{person}{Chengsheng Mao}, {and} \bibinfo{person}{Yuan Luo}.} \bibinfo{year}{2019}\natexlab{}.
\newblock \showarticletitle{KG-BERT: BERT for Knowledge Graph Completion}.
\newblock \bibinfo{journal}{\emph{arXiv preprint arXiv:1909.03193}} (\bibinfo{year}{2019}).
\newblock


\bibitem[Zhai et~al\mbox{.}(2022)]%
        {lit}
\bibfield{author}{\bibinfo{person}{Xiaohua Zhai}, \bibinfo{person}{Xiao Wang}, \bibinfo{person}{Basil Mustafa}, \bibinfo{person}{Andreas Steiner}, \bibinfo{person}{Daniel Keysers}, \bibinfo{person}{Alexander Kolesnikov}, {and} \bibinfo{person}{Lucas Beyer}.} \bibinfo{year}{2022}\natexlab{}.
\newblock \showarticletitle{LiT: Zero-Shot Transfer with Locked-image text Tuning}. In \bibinfo{booktitle}{\emph{2022 IEEE/CVF Conference on Computer Vision and Pattern Recognition (CVPR)}}. \bibinfo{pages}{18102--18112}.
\newblock
\href{https://doi.org/10.1109/CVPR52688.2022.01759}{doi:\nolinkurl{10.1109/CVPR52688.2022.01759}}


\bibitem[Zhang et~al\mbox{.}(2023)]%
        {mart}
\bibfield{author}{\bibinfo{person}{Ningyu Zhang}, \bibinfo{person}{Lei Li}, \bibinfo{person}{Xiang Chen}, \bibinfo{person}{Xiaozhuan Liang}, \bibinfo{person}{Shumin Deng}, {and} \bibinfo{person}{Huajun Chen}.} \bibinfo{year}{2023}\natexlab{}.
\newblock \showarticletitle{Multimodal Analogical Reasoning over Knowledge Graphs}. In \bibinfo{booktitle}{\emph{The Eleventh International Conference on Learning Representations}}.
\newblock
\urldef\tempurl%
\url{https://openreview.net/forum?id=NRHajbzg8y0P}
\showURL{%
\tempurl}


\bibitem[Zhang et~al\mbox{.}(2020)]%
        {mmrfan}
\bibfield{author}{\bibinfo{person}{Yingying Zhang}, \bibinfo{person}{Quan Fang}, \bibinfo{person}{Shengsheng Qian}, {and} \bibinfo{person}{Changsheng Xu}.} \bibinfo{year}{2020}\natexlab{}.
\newblock \showarticletitle{Multi-modal Multi-relational Feature Aggregation Network for Medical Knowledge Representation Learning}. In \bibinfo{booktitle}{\emph{Proceedings of the 28th ACM International Conference on Multimedia}} (Seattle, WA, USA) \emph{(\bibinfo{series}{MM '20})}. \bibinfo{publisher}{Association for Computing Machinery}, \bibinfo{address}{New York, NY, USA}, \bibinfo{pages}{3956–3965}.
\newblock
\showISBNx{9781450379885}
\href{https://doi.org/10.1145/3394171.3413736}{doi:\nolinkurl{10.1145/3394171.3413736}}


\bibitem[Zhao et~al\mbox{.}(2022)]%
        {mose}
\bibfield{author}{\bibinfo{person}{Yu Zhao}, \bibinfo{person}{Xiangrui Cai}, \bibinfo{person}{Yike Wu}, \bibinfo{person}{Haiwei Zhang}, \bibinfo{person}{Ying Zhang}, \bibinfo{person}{Guoqing Zhao}, {and} \bibinfo{person}{Ning Jiang}.} \bibinfo{year}{2022}\natexlab{}.
\newblock \showarticletitle{{M}o{SE}: Modality Split and Ensemble for Multimodal Knowledge Graph Completion}. In \bibinfo{booktitle}{\emph{Proceedings of the 2022 Conference on Empirical Methods in Natural Language Processing}}, \bibfield{editor}{\bibinfo{person}{Yoav Goldberg}, \bibinfo{person}{Zornitsa Kozareva}, {and} \bibinfo{person}{Yue Zhang}} (Eds.). \bibinfo{publisher}{Association for Computational Linguistics}, \bibinfo{address}{Abu Dhabi, United Arab Emirates}, \bibinfo{pages}{10527--10536}.
\newblock
\href{https://doi.org/10.18653/v1/2022.emnlp-main.719}{doi:\nolinkurl{10.18653/v1/2022.emnlp-main.719}}


\bibitem[Zhu et~al\mbox{.}(2023)]%
        {fedlu}
\bibfield{author}{\bibinfo{person}{Xiangrong Zhu}, \bibinfo{person}{Guangyao Li}, {and} \bibinfo{person}{Wei Hu}.} \bibinfo{year}{2023}\natexlab{}.
\newblock \showarticletitle{Heterogeneous Federated Knowledge Graph Embedding Learning and Unlearning}. In \bibinfo{booktitle}{\emph{Proceedings of the ACM Web Conference 2023}} (Austin, TX, USA) \emph{(\bibinfo{series}{WWW '23})}. \bibinfo{publisher}{Association for Computing Machinery}, \bibinfo{address}{New York, NY, USA}, \bibinfo{pages}{2444–2454}.
\newblock
\showISBNx{9781450394161}
\href{https://doi.org/10.1145/3543507.3583305}{doi:\nolinkurl{10.1145/3543507.3583305}}


\bibitem[Ziegler et~al\mbox{.}(2005)]%
        {ild_measure}
\bibfield{author}{\bibinfo{person}{Cai-Nicolas Ziegler}, \bibinfo{person}{Sean~M. McNee}, \bibinfo{person}{Joseph~A. Konstan}, {and} \bibinfo{person}{Georg Lausen}.} \bibinfo{year}{2005}\natexlab{}.
\newblock \showarticletitle{Improving recommendation lists through topic diversification}. In \bibinfo{booktitle}{\emph{Proceedings of the 14th International Conference on World Wide Web}} (Chiba, Japan) \emph{(\bibinfo{series}{WWW '05})}. \bibinfo{publisher}{Association for Computing Machinery}, \bibinfo{address}{New York, NY, USA}, \bibinfo{pages}{22–32}.
\newblock
\showISBNx{1595930469}
\href{https://doi.org/10.1145/1060745.1060754}{doi:\nolinkurl{10.1145/1060745.1060754}}


\end{thebibliography}

\clearpage

\nobalance

\appendix

\pagestyle{supplementarystyle}

\section{Supplementary Material}\label{sec:supplementary_material}

The supplementary material\protect\footnote{This supplementary material extends the version published in the Proceedings of the ACM Web Conference 2026 (WWW ’26). Additional details originally omitted due to the conference page limit are included here.} provides additional analyses and methodological details that complement the main paper, including: (i) a description of WikiArt-v2 dataset collection and WikiArt-MKG-v2 knowledge graph construction (Section~\ref{app:dataset_curation}); (ii) evaluation of \textit{isRelatedTo} relations on WikiArt-MKG-v2 (Section~\ref{app:related_to_evaluation}); (iii) linear-probe evaluation of pretrained visual encoders for artwork classification on WikiArt-v1 and WikiArt-v2 (Section~\ref{app:linear_probe}); (iv) ablations on modality fine-tuning on WikiArt-MKG-v1 and WikiArt-MKG-v2 (Section~\ref{app:modality_finetuning}); (v) comparison of fusion strategies on WN9-IMG (Section~\ref{app:wn9_ablation}); (vi) per-relation performance analysis on WikiArt-MKG-v2 (Section~\ref{app:mrr_per_relation}); (vii) discussion of limitations (Section~\ref{app:limitations}); and (viii) broader impact considerations (Section~\ref{app:broader_impact}).

\subsection{WikiArt-v2 Dataset Collection}\label{app:dataset_curation}

WikiArt-v2 was constructed through large-scale web scraping of the WikiArt online collection~\cite{wikiart}, a comprehensive repository of fine art spanning multiple centuries, artistic movements, and geographic regions. The dataset extends WikiArt-v1~\cite{tan_artgan, elgammal2018shape, artsagenet} both in scale and semantic richness, increasing coverage from 76K to 217K artworks and from 750 to 4.2K artists.

\begin{table}[t]
\centering
\caption{Number of unique head and tail entities per relation in WikiArt-MKG-v2. The top section shows artwork-to-attribute, artist-to-attribute, and artist-to-artist relations, while the bottom section shows the two relatedness relations.}
\label{tab:wikiart_mkg_v2_statistics}
\resizebox{\columnwidth}{!}{\begin{tabular}{@{} l c r}\toprule
{Head} & {Relation} & {Tail} \\ \midrule
143,102 & isAssociatedWithTag & 1,937\\
212,649 & isCreatedByArtist & 4,201\\
197,335 & belongsToGenre & 67\\
158,160 & isCreatedInYear & 620\\
81,277 & isCreatedWithMedium & 149\\
134,939 & hasStyle & 20\\
42,190 & isLocatedIn & 387\\
3,693 & isAssociatedWithField & 21\\
4,004 & hasNationality & 49\\
2,803 & isAssociatedWithArtMovement & 86\\
4,149 & wasBornOnDate & 13\\
3,172 & diedOnDate & 13\\
1,212 & diedIn & 23\\
1,179 & wasBornIn & 23\\
664 & isMemberOfPaintingSchool & 43\\
534 & isAffiliatedWithArtInstitution & 26\\
287 & isInfluencedBy & 218\\
218 & isInfluencedOn & 286\\
111 & isTeacherOf & 146\\
145 & isPupilOf & 110\\ \midrule
200,699 & isRelatedToArtwork & 131,274\\
3,479 & isRelatedToArtist & 3,101\\
\bottomrule
\end{tabular}}
\end{table}

\textbf{Data collection and curation.} We implemented a multi-stage scraping pipeline to collect artwork images, artist profiles, and associated metadata. For each artwork, we extracted high-resolution images, basic metadata (e.g., title, year, dimensions), categorical attributes (e.g., style, genre, medium), and relational metadata. For artists, we collected biographical information, movement affiliations, and inter-artist relationships. To ensure data quality, we applied several filtering criteria. Artworks with missing images or critical metadata (e.g., artist, creation year) were excluded. Entity names were normalized to resolve aliases and spelling variations, particularly for artists and locations appearing in multiple languages. Artist relationships were retained only when explicitly documented in structured metadata fields. This combination of scale, diversity, and rich metadata makes WikiArt-v2 a comprehensive resource for computational fine art analysis.

\begin{table}[t]
\centering
\caption{Results for the \textit{isRelatedToArtwork} relation. We report average precision (AP) for each attribute, mean AP (mAP), and intra-list diversity (ILD) based on attributes (ILD-A) and visual features (ILD-V). Best results for AP and mAP are in \textbf{bold}, second best \underline{underlined}.}
\label{tab:wikiart_v2_related_artworks}
\resizebox{\columnwidth}{!}{\begin{tabular}{@{} lccccccc @{}}
\toprule
Method & \multicolumn{4}{c}{AP} & mAP & ILD-A & ILD-V \\
\cmidrule(lr){2-5} 
 & Style & Genre & Tag & Date & & &  \\
\midrule
Zero-shot CLIP & 0.521 & \textbf{0.615} & \textbf{0.299} & 0.579 & \underline{0.503} & 0.638 & 0.767 \\
VL-TransE    & \underline{0.523} & \underline{0.613} & \textbf{0.299} & 0.581 & \textbf{0.504} & 0.638 & 0.768 \\
VL-DistMult    & \textbf{0.539} & 0.601 & 0.283 & \textbf{0.592} & \textbf{0.504} & 0.645 & 0.779  \\
VL-ComplEx    & \textbf{0.539} & 0.602 & \underline{0.285} & \textbf{0.592} & \textbf{0.504} & 0.645 & 0.779 \\
VL-RotatE    & 0.445 & 0.594 & 0.277 & 0.505 & 0.455 & 0.666 & 0.761 \\
\bottomrule
\end{tabular}}
\end{table}

\textbf{WikiArt-MKG-v2 construction.} WikiArt-MKG-v2 is a multimodal knowledge graph derived from WikiArt-v2 and contains 22 relation types spanning four categories: artwork-to-attribute, artist-to-attribute, artwork-to-artwork, and artist-to-artist relations. Table~\ref{tab:wikiart_mkg_v2_statistics} reports the number of unique head and tail entities for each relation type. Artwork-to-attribute relations connect artworks to creation context (\textit{isCreatedByArtist}, \textit{isCreatedInYear}, \textit{isCreatedWithMedium}), stylistic properties (\textit{hasStyle}, \textit{belongsToGenre}, \textit{isAssociatedWithTag}), and spatial context (\textit{isLocatedIn}). Artist-to-attribute relations model biographical information (\textit{wasBornOnDate}, \textit{diedOnDate}, \textit{wasBornIn}, \textit{diedIn}, \textit{hasNationality}), professional affiliations (\textit{isAssociatedWithField}, \textit{isAssociatedWithArtMovement}, \textit{isMemberOfPaintingSchool}, \textit{isAffiliatedWithArtInstitution}), and interpersonal relationships (\textit{isInfluencedBy}, \textit{isInfluencedOn}, \textit{isTeacherOf}, \textit{isPupilOf}). The two relatedness relations (\textit{isRelatedToArtwork}, \textit{isRelatedToArtist}) capture semantic similarity beyond explicit categorical attributes and are discussed separately in Section~\ref{app:related_to_evaluation}. For each artwork-attribute pair in the scraped metadata, we generated a corresponding triple. Artist-to-artist relations (e.g., \textit{isInfluencedBy}, \textit{isPupilOf}) were extracted from structured relationship fields in artist profiles. For consistency, we validated inverse relations and removed contradictory or duplicate triples. The relatedness relations were constructed using similarity annotations provided by the WikiArt online collection~\cite{wikiart}, which define related artworks and artists based on shared stylistic, temporal, and thematic properties. To reduce noise from rare entities, we applied frequency-based filtering to most attribute types, omitting attributes that occur fewer than 10 times in the dataset. This threshold effectively removes highly specific or potentially erroneous labels while retaining sufficient diversity in categorical attributes such as styles, genres, tags, and locations. We exempted the inductive artist-to-artist relations (\textit{isInfluencedBy}, \textit{isInfluencedOn}, \textit{isTeacherOf}, \textit{isPupilOf}) from this filtering, as these relations are inherently sparse and their low frequency reflects the specialized nature of artistic relationships. For temporal attributes, we applied coarse-grained binning to reduce sparsity. Artist birth and death dates (\textit{wasBornOnDate}, \textit{diedOnDate}) were transformed into half-century periods (e.g., 1600--1649, 1650--1699), reflecting typical granularity in art historical analysis. Similarly, for geographic attributes (\textit{wasBornIn}, \textit{diedIn}), we mapped fine-grained place names to their corresponding countries, reducing the excessive fragmentation caused by city-level annotations while still preserving meaningful national-level geographic distinctions.

\textbf{Inductive split construction.} A critical design choice in WikiArt-MKG-v2 is how we construct the train/validation/test splits for systematic evaluation of inductive inference. All test artworks are unseen during training, requiring the model to generalize to new visual entities. For artist-to-artist relations (e.g., \textit{isInfluencedBy}, \textit{isPupilOf}), we use disjoint artist subsets: triples connecting training and test artists are removed entirely, ensuring that the model performs inductive reasoning over artist relationships. For artwork-to-artist relations, artists appear in the training set through connections to other artworks, but their specific connections to test artworks remain unseen. This setup reflects realistic deployment scenarios, such as museum collections, where new artworks are continuously added and their attributes must be inferred without retraining.

\textbf{WikiArt-MKG-v1 vs WikiArt-MKG-v2.} WikiArt-MKG-v1, a multimodal knowledge graph derived from WikiArt-v1~\cite{tan_artgan, elgammal2018shape, artsagenet}, serves as a focused benchmark for artwork-attribute prediction, consisting of 76K artworks connected to artists, styles, creation years, and tags through 4 relation types in a modality-asymmetric setting. WikiArt-MKG-v2 substantially extends WikiArt-MKG-v1 along three dimensions: (i) scale, with nearly three times more artworks and artists, as well as richer metadata; (ii) structural diversity, expanding to 22 relation types, including artist-to-artist relations for modeling influence networks; and (iii) relational heterogeneity, with cardinalities ranging from sparse relations (e.g., \textit{isPupilOf}, \textit{isTeacherOf}) to broad relations (e.g., \textit{isCreatedByArtist}, \textit{belongsToGenre}). Unlike curated benchmarks such as WN9-IMG~\cite{wn9_img_dataset}, where all entities possess complete modalities by design, both WikiArt-MKGs exhibit modality sparsity, providing a realistic setting for evaluating multimodal KGE methods under modality asymmetry.

\begin{table}[t]
\centering
\caption{Results for the \textit{isRelatedToArtist} relation. We report average precision (AP) for each attribute, mean AP (mAP), and intra-list diversity (ILD) based on attributes (ILD-A) and textual features (ILD-T). Best results for AP and mAP are in \textbf{bold}, second best \underline{underlined}.}
\label{tab:wikiart_v2_related_artists}
\resizebox{\columnwidth}{!}{\begin{tabular}{@{} lcccccc @{}}
\toprule
Method & \multicolumn{3}{c}{AP} & mAP & ILD-A & ILD-T \\
\cmidrule(lr){2-4} 
 & Movement & School  & Date & & &  \\
\midrule
Zero-shot CLIP & 0.292 & 0.217 & 0.524 & 0.344 & 0.845 & 0.462 \\
VL-TransE    & \textbf{0.486} & \textbf{0.445} & \textbf{0.782} & \textbf{0.571} & 0.702 & 0.466  \\
VL-DistMult    & 0.356 & 0.334 & 0.695 & 0.462 & 0.794 & 0.438 \\
VL-ComplEx    & 0.336 & 0.278 & 0.673 & 0.429 & 0.822 & 0.442 \\
VL-RotatE    & \underline{0.440} & \underline{0.409} & \underline{0.729} & \underline{0.526} & 0.728 & 0.467 \\
\bottomrule
\end{tabular}}
\end{table}

\subsection{Evaluation of \textit{isRelatedTo} Relations}\label{app:related_to_evaluation}

The two relatedness relations (\textit{isRelatedToArtwork}, \textit{isRelatedToArtist}) differ fundamentally from other relations in WikiArt-MKG-v2 in both their semantics and the nature of their ground truth, necessitating a distinct evaluation protocol.

\textbf{Ground truth limitations.} Unlike categorical relations such as \textit{isCreatedByArtist} or \textit{hasStyle}, where ground truth is unambiguous and complete, relatedness relations capture subjective notions of semantic similarity. The relatedness annotations provided by the WikiArt online collection~\cite{wikiart} specify related entities for each query entity, but these annotations are inherently incomplete, as many entities sharing stylistic, temporal, or thematic properties may not be annotated as related. Consequently, standard link prediction metrics such as MRR and Hits@K would unfairly penalize models for retrieving valid but unannotated related entities.

\textbf{Experimental setup.} We adopt a retrieval-based evaluation framework assessing both precision and diversity. We measure average precision (AP) per attribute category and mean AP (mAP) as relevance metrics, alongside intra-list diversity (ILD)~\cite{ild_measure} to assess the quality of the predicted recommendations. We define relevance based on shared attributes following the relatedness criteria defined by WikiArt: style, period, genre, and tag for artworks; movement, painting school, and birth date for artists. This attribute-based definition provides a principled and reproducible relevance judgment that enables systematic evaluation across large-scale test sets. We retrieve the top-50 predicted entities for artworks and the top-20 for artists. For baselines, we use CLIP visual embeddings with cosine similarity for artworks and textual prompts for artists. ILD is computed over both continuous embeddings and discrete attributes. For artworks, we measure diversity using ViT-based visual features (ILD-V) and categorical attributes (ILD-A), whereas for artists, we use BERT-based textual features (ILD-T) and categorical attributes (ILD-A). The interpretation varies by context: for artworks with a large candidate pool, higher ILD indicates desirable exploration; for artists with a smaller pool targeting specific relationships, lower ILD may reflect semantically coherent retrieval.

\textbf{Results and discussion.} Tables~\ref{tab:wikiart_v2_related_artworks} and~\ref{tab:wikiart_v2_related_artists} show that VL-KGE models outperform zero-shot CLIP in precision while maintaining competitive diversity. For artworks, the best VL-KGE variants attain the highest precision with strong diversity across visual and attribute measures. For artists, VL-KGE variants consistently outperform zero-shot CLIP. The qualitative analysis in Figure~\ref{fig:qualitative_analysis_clip_complex} of the main paper provides additional insights. For the \textit{isRelatedToArtwork} relation, zero-shot CLIP tends to retrieve artworks that are visually similar based on low-level perceptual features such as color palette, composition, and subject matter. In contrast, VL-ComplEx retrieves related artworks that share deeper semantic characteristics with the query. For example, given Rembrandt's \textit{Flora} (1634) as a query, VL-ComplEx retrieves two self-portraits by Rembrandt---artworks that may not appear immediately visually similar to \textit{Flora} but share the critical attribute of being created by the same artist, demonstrating the model's ability to capture artist-specific stylistic patterns and thematic continuity beyond surface-level visual similarity.

\begin{table}[t]
\centering
\caption{Linear-probe results on WikiArt-v1 and WikiArt-v2. Style, Artist, and Timeframe report accuracy; Tags reports mAP; and Date$\ddagger$ reports cumulative accuracy with $\theta=10$ ($\pm$10 years). For all metrics, higher is better. Best results are in \textbf{bold}, second best \underline{underlined}.}
\label{tab:linear_probe_results}
\resizebox{\columnwidth}{!}{\begin{tabular}{@{} l *{4}{c}  *{4}{c}  @{}}
\toprule Method & \multicolumn{4}{c}{WikiArt-v1} & \multicolumn{4}{c}{WikiArt-v2} \\
\cmidrule(lr){2-5} \cmidrule(l){6-9}
& Style  & Artist  & Timeframe  & Tags & Style  & Artist  & Date$\ddagger$  & Tags  \\
    \midrule
    ViT   & 0.560 & \underline{0.445} & 0.599 & 0.576
          & 0.514 & 0.395 & 0.104 & 0.174 \\
    BLIP  & \underline{0.611} & 0.412 & \underline{0.655} & \underline{0.611}
          & \underline{0.586} & \underline{0.430} & \underline{0.142} & \underline{0.280} \\
    CLIP  & \textbf{0.737} & \textbf{0.701} & \textbf{0.773} & \textbf{0.690}
          & \textbf{0.698} & \textbf{0.647} & \textbf{0.153} & \textbf{0.364} \\
    \bottomrule
\end{tabular}}
\end{table}

\subsection{Linear-Probe Evaluation}\label{app:linear_probe}

\textbf{Evaluation protocol.} We evaluate the quality of pretrained visual encoders via linear probing on WikiArt-v1 and WikiArt-v2, freezing the encoders and training only a linear classifier. Following~\cite{artsagenet}, we evaluate four tasks: style classification, artist attribution, timeframe estimation, and multi-label tag prediction. For WikiArt-v2, we report cumulative year prediction accuracy (Date$\ddagger$) with a $\theta=10$ year threshold, measuring whether the predicted creation year falls within $\pm$ 10 years of ground truth. WikiArt-v2's scale (217K artworks and 4.2K artists, compared to 76K artworks and 750 artists in WikiArt-v1) presents increased difficulty, particularly for artist attribution and creation year estimation.

\textbf{Results and encoder comparison.} Table~\ref{tab:linear_probe_results} shows that CLIP consistently outperforms ViT and BLIP across all settings. CLIP’s contrastive pretraining produces visual representations that transfer effectively to fine art categorization tasks. ViT’s ImageNet pretraining shows limited transfer to artistic images, while BLIP’s generative pretraining objective underperforms contrastive pretraining on discriminative downstream tasks in the fine art domain.

\subsection{Ablation on Modality Fine-Tuning}\label{app:modality_finetuning}

We investigate the effect of fine-tuning pretrained vision-language encoders in VL-ComplEx on WikiArt-MKGs (Table~\ref{tab:wikiart_v1_v2_ablation}). We consider four configurations: freezing both encoders, fine-tuning the visual encoder, the textual encoder, or both. On WikiArt-MKG-v1, fine-tuning the textual encoder performs best, as structural embeddings are not used in this setting, and fine-tuning the textual representations aligns them better with the KG domain. Conversely, for WikiArt-MKG-v2, freezing both encoders leads to optimal results, as structural embeddings already capture relational structure.

\subsection{Ablation on Fusion Strategies}\label{app:wn9_ablation}

Table~\ref{tab:wn9_img_dataset_ablation} reports results of CLIP-based VL-DistMult on the WN9-IMG validation set under different modality configurations and fusion strategies. Incorporating either visual or textual information improves performance over the unimodal baseline, with visual features contributing slightly more, which is consistent with WN9-IMG’s origin from ImageNet synsets. Combining both modalities yields the best results overall. Among fusion methods, average fusion performs best, while weighted fusion remains competitive.

\begin{table}[t]
\centering
\caption{Results using VL-ComplEx (base: CLIP) on the WikiArt-MKG-v1 and WikiArt-MKG-v2 validation sets for fine-tuning different modalities. For all metrics, higher is better. Best results are in \textbf{bold}, second best \underline{underlined}.}
\label{tab:wikiart_v1_v2_ablation}
\resizebox{\columnwidth}{!}{
\begin{tabular}{@{} cc cccc cccc @{}}
\toprule
\multicolumn{2}{c}{Fine-tuning}
& \multicolumn{4}{c}{WikiArt-MKG-v1} 
& \multicolumn{4}{c}{WikiArt-MKG-v2} \\
\cmidrule(r){1-2} \cmidrule(lr){3-6} \cmidrule(l){7-10}
 Visual & Textual & MRR & Hits@1 & Hits@3 & Hits@10 
    & MRR & H@1 & H@3 & H@10 \\
\midrule
\xmark & \xmark      & \underline{0.784} & \underline{0.661} & \underline{0.892} & \textbf{0.974} 
                         &  \textbf{0.577} & \textbf{0.463} & \textbf{0.642} & \textbf{0.794} \\
\checkmark & \xmark  & 0.753 & 0.630 & 0.854 & \underline{0.959} 
                         & 0.566 & 0.453 & 0.631 & 0.783  \\
\xmark & \checkmark   & \textbf{0.787} & \textbf{0.667} & \textbf{0.893} & \textbf{0.974} 
                         & \underline{0.575} & \underline{0.461} & \underline{0.641} & \underline{0.792}  \\
\checkmark & \checkmark & 0.753 & 0.632 & 0.852 & 0.958 
                         & 0.566 & 0.453 & 0.629 & 0.781 \\
\bottomrule
\end{tabular}
}
\end{table}

\begin{table*}[t]
\centering
\caption{Results using VL-DistMult (base: CLIP) on the WN9-IMG validation set under different modality configurations and fusion strategies. For all metrics, higher is better. Best results are in \textbf{bold}, second best \underline{underlined}.}
\label{tab:wn9_img_dataset_ablation}
\resizebox{\textwidth}{!}{%
\begin{tabular}{@{} cc *{4}{c} *{4}{c} *{4}{c} @{}}
\toprule
\multicolumn{2}{c}{Modality} &
\multicolumn{4}{c}{Average} &
\multicolumn{4}{c}{Concatenation} &
\multicolumn{4}{c}{Weighted} \\
\cmidrule(r){1-2}\cmidrule(lr){3-6}\cmidrule(lr){7-10}\cmidrule(l){11-14}
Visual & Textual &
MRR & Hits@1 & Hits@3 & Hits@10 &
MRR & Hits@1 & Hits@3 & Hits@10 &
MRR & Hits@1 & Hits@3 & Hits@10 \\
\midrule

\xmark & \xmark &
 0.898 & 0.896 & 0.898 & 0.899 &
 0.898 & 0.896 & 0.898 & 0.899 &
 0.898 & 0.896 & 0.898 & 0.899 \\
\cmark & \xmark &
0.921 & \underline{0.913} & 0.924 & \underline{0.939} &
0.910 & 0.905 & 0.909 & 0.918 &
0.921 & 0.912 & \underline{0.925} & 0.938 \\

\xmark & \cmark &
0.919 & 0.909 & 0.923 & 0.938 &
0.909 & 0.905 & 0.909 & 0.917 &
0.919 & 0.910 & 0.923 & 0.936 \\

\cmark & \cmark &
\textbf{0.924} & \underline{0.913} & \textbf{0.927} & \textbf{0.950} &
0.907 & 0.905 & 0.907 & 0.909 &
\underline{0.922} & \textbf{0.914} & 0.924 & \underline{0.939} \\

\bottomrule
\end{tabular}
}
\end{table*}

\begin{figure*}[t]
    \centering
    \includegraphics[width=\textwidth]{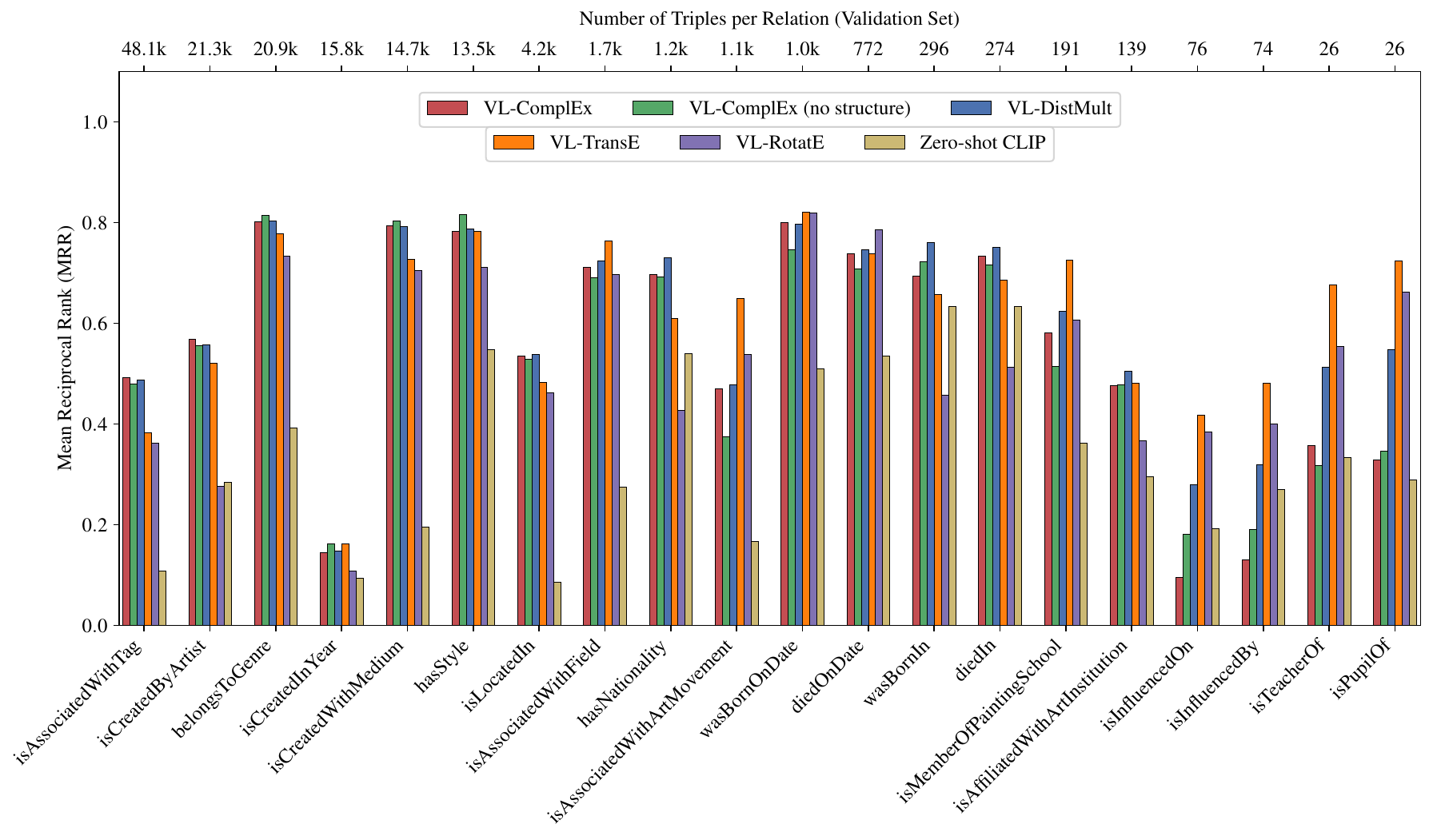}
    \caption{Per-relation mean reciprocal rank (MRR) on the WikiArt-MKG-v2 validation set for zero-shot CLIP and VL-KGEs.}
    \label{fig:mrr_per_relation}
\end{figure*}

\subsection{Per-Relation Analysis on WikiArt-MKG-v2}\label{app:mrr_per_relation}

Figure~\ref{fig:mrr_per_relation} reports per-relation mean reciprocal rank (MRR) for the best-performing VL-KGE variants on the WikiArt-MKG-v2 validation set. Results show substantial variability across relation types, reflecting interactions between semantics and model architecture. Bilinear models (DistMult, ComplEx) excel at many-to-one relations with abundant training data (e.g., \textit{isCreatedByArtist}, \textit{isAssociatedWithTag}). In contrast, translational models (TransE, RotatE) excel at sparse, inverse artist-to-artist relations (e.g., \textit{isInfluencedBy}, \textit{isInfluencedOn}), which require directional reasoning that these models naturally encode through translation and rotation operations. VL-ComplEx (no structure) performs comparably using only VLM features, highlighting the representational power of pretrained VLMs.

\subsection{Limitations}\label{app:limitations}

While VL-KGE demonstrates strong performance across multiple benchmarks, several limitations should be noted. First, our framework encodes all attributes as text, which may be suboptimal for temporal or geospatial properties that could benefit from specialized numerical or coordinate-based representations. Second, the quality of multimodal representations depends on the pretrained vision-language encoders, which may exhibit domain-specific performance variations or encode patterns from their pretraining data that do not fully align with downstream tasks. Third, WikiArt-MKG-v2, while substantially larger than existing fine art benchmarks, is constructed from a single data source and may not comprehensively capture the full breadth of art historical scholarship. Fourth, our empirical evaluation focuses primarily on fine art knowledge graphs; broader assessment across additional domains exhibiting modality asymmetry would further validate generalizability. Finally, while VL-KGE enables inductive inference over unseen entities through pretrained encoders, relation embeddings still require training on observed triples, which may limit applicability in scenarios with extremely sparse supervision or entirely novel relation types.

\subsection{Broader Impact}\label{app:broader_impact}

This work contributes to advancing multimodal knowledge graph representations, with potential applications in digital humanities, cultural heritage, and information retrieval. By improving link prediction in knowledge graphs with modality asymmetry, VL-KGE may enhance computational tools for art historical research, collection management, and educational platforms. The framework could assist scholars and practitioners in discovering stylistic connections, analyzing artistic influence networks, and contextualizing cultural artifacts within structured knowledge graphs. As with any knowledge graph-based system, practitioners should remain cognizant that the underlying data sources may reflect particular cultural, historical, or institutional perspectives. Future work should prioritize expansion to more diverse cultural collections and non-Western art traditions to ensure broader and more inclusive representational coverage. Responsible deployment requires awareness of potential representational gaps and careful consideration of how automated systems might be integrated with expert human judgment in curatorial and educational contexts. Such systems should therefore be used as assistive tools that support exploration and discovery, rather than as authoritative arbiters of artistic interpretation.

\end{document}